\documentclass[sigconf,nonacm]{acmart}
\usepackage{multirow}
\usepackage{subfigure}
\usepackage{blindtext}
\usepackage{enumitem}
\usepackage{algorithm}  
\usepackage{algpseudocode}  
\usepackage{amsmath}  
\usepackage{hyperref}
\usepackage{flushend}
\usepackage{balance}
\usepackage{diagbox}
\usepackage{colortbl}
\usepackage{xcolor}
\usepackage{array} 
\definecolor{o1}{HTML}{FFEDE0} 
\definecolor{o2}{HTML}{FFDFC8} 
\definecolor{o3}{HTML}{FFD6B7} 
\definecolor{o4}{HTML}{FFCAA3} 
\definecolor{o5}{HTML}{FFBE8F} 
\definecolor{o6}{HTML}{FFB37C} 
\definecolor{o7}{HTML}{FFA665} 
\definecolor{o8}{HTML}{FF994F} 
\definecolor{o9}{HTML}{FF7D1F} 
\definecolor{o10}{HTML}{FF6B00} 

\renewcommand{\algorithmicrequire}{\textbf{Input:}}  
\renewcommand{\algorithmicensure}{\textbf{Output:}} 

\AtBeginDocument{%
\providecommand\BibTeX{{%
  \normalfont B\kern-0.5em{\scshape i\kern-0.25em b}\kern-0.8em\TeX}}}

  \copyrightyear{2022} 
  \acmYear{2022} 
  \setcopyright{acmcopyright}
  \acmConference[CIKM '22]{Proceedings of the 31st ACM International Conference on Information and Knowledge Management}{October 17--21, 2022}{Atlanta, GA, USA.}
  \acmBooktitle{Proceedings of the 31st ACM International Conference on Information and Knowledge Management (CIKM '22), October 17--21, 2022, Atlanta, GA, USA}
  \acmPrice{15.00}
  \acmDOI{10.1145/xxxxxx.xxxxxx}
  \acmISBN{978-1-4503-9236-5/22/10}
\begin{document}

\title{Learning List-wise Representation in Reinforcement Learning for Ads Allocation with Multiple Auxiliary Tasks
}



\author{Ze Wang}
\authornote{Equal contribution. Listing order is random.}
\affiliation{%
 \institution{Meituan}
 \city{Beijing}
 \country{China}
}
\email{wangze18@meituan.com}

\author{Guogang Liao}
\authornotemark[1]
\affiliation{%
 \institution{Meituan}
 \city{Beijing}
 \country{China}
}
\email{liaoguogang@meituan.com}

\author{Xiaowen Shi}
\authornote{Corresponding author.}
\affiliation{%
 \institution{Meituan}
 \city{Beijing}
 \country{China}
}
\email{shixiaowen03@meituan.com}

\author{Xiaoxu Wu}
\affiliation{%
 \institution{Meituan}
 \city{Beijing}
 \country{China}
}
\email{wuxiaoxu04@meituan.com}

\author{Chuheng Zhang}
\authornote{This work was done when Chuheng Zhang was an intern in Meituan.}
\affiliation{%
 \institution{Meituan}
 \city{Beijing}
 \country{China}
}
\email{zhangchuheng123@live.com}

\author{Yongkang Wang}
\affiliation{%
 \institution{Meituan}
 \city{Beijing}
 \country{China}
}
\email{wangyongkang03@meituan.com}

\author{Xingxing Wang}
\affiliation{%
 \institution{Meituan}
 \city{Beijing}
 \country{China}
}
\email{wangxingxing04@meituan.com}

\author{Dong Wang}
\affiliation{%
 \institution{Meituan}
 \city{Beijing}
 \country{China}
}
\email{wangdong07@meituan.com}
\renewcommand{\shortauthors}{Ze Wang and Guogang Liao, et al.}

\begin{abstract}
  With the recent prevalence of reinforcement learning (RL), there have been tremendous interests in utilizing RL for ads allocation in recommendation platforms (e.g., e-commerce and news feed sites). 
  To achieve better allocation, the input of recent RL-based ads allocation methods is upgraded from point-wise single item to list-wise item arrangement.
  However, this also results in a high-dimensional space of state-action pairs, making it difficult to learn list-wise representations with good generalization ability. This further hinders the exploration of RL agents and causes poor sample efficiency. 
  To address this problem, we propose a novel RL-based approach for ads allocation which learns better list-wise representations by leveraging task-specific signals on Meituan food delivery platform. 
  Specifically, we propose three different auxiliary tasks based on reconstruction, prediction, and contrastive learning respectively according to prior domain knowledge on ads allocation.
  We conduct extensive experiments on Meituan food delivery platform to evaluate the effectiveness of the proposed auxiliary tasks. Both offline and online experimental results show that the proposed method can learn better list-wise representations and achieve higher revenue for the platform compared to the state-of-the-art baselines.
\end{abstract}


\maketitle

\section{Introduction}
Ads and organic items are mixed together and displayed to users in e-commerce feed nowadays \cite{Ghose2009AnEA}. E-commerce platforms gain the platform service fee (hereinafter referred to as fee) according to orders and charge advertisers based on exposures or clicks. 
The increase number of displayed ads brings higher ads revenue, but worse user experience, which results in a decrease in order quantify and fee \cite{Zhao2018ImpressionAF}. Therefore, how to allocate limited ad slots effectively to maximize the overall revenue (i.e., fee and ads revenue) has been considered a meaningful and challenging problem \cite{Wang2011LearningTA,Mehta2013OnlineMA,zhang2018whole}. 
Unlike the practice of allocating ads to pre-determined slots \cite{ouyang2020minet,li2020deep,Carrion2021BlendingAW, Wei2020GeneratorAC, Feng2021RevisitRS}, recent dynamic ads allocation strategies usually model the ads allocation problem as an Markov Decision Process (MDP) \cite{sutton1998introduction} and solve it using reinforcement learning (RL) \cite{zhang2018whole,liao2021cross,zhao2019deep, Feng2018LearningTC, zhao2020jointly}. 
For instance,
\citet{xie2021hierarchical} propose a hierarchical RL-based framework which first decides on the type of the item to present and then determines the specific item for each slot.
However, this work make decisions based on the point-wise representation of candidate items, without considering the crucial arrangement signal \cite{liao2021cross} hidden in item arrangement. 
\citet{zhao2019deep} and \citet{liao2021cross} propose different DQN architectures to achieve better performance, which both take the list-wise representation of item arrangement as input and allocate the slots in one screen at a time.
However, these algorithms encounter one major challenge:
 The rarity of the list-wise item arrangement results in a high-dimensional state-action space. For example, the number of candidate items in each slot on Meituan food delivery platform is more than millions, which leads to the curse of dimensionality for item arrangements in multiple slots of one screen. The high-dimensional state-action space makes it difficult to learn a generalizable list-wise representation, which further causes poor sample efficiency and suboptimal performance. 

Utilizing auxiliary task is one common solution for representation learning in RL \cite{hafner2019dream, hafner2019learning, guo2020bootstrap,mazoure2020deep,rcrl}. For instance, \citet{finn2016deep} present an auxiliary task for learning state representation using deep spatial autoencoders. \citet{jaderberg2016reinforcement} propose an auxiliary task to predict the onset of immediate reward given some historical context. \citet{rcrl} leverage return to construct a contrastive auxiliary task for speeding up the main RL task. 
However, domain knowledge is very helpful information that has not been fully explored. Most existing auxiliary tasks for representation learning lack the utilization of domain knowledge in e-commerce scenario, which makes them unable to achieve good performance on ads allocation.

To this end, we propose an auxiliary-task based RL method, which aims to learn an efficient representation by leveraging task-specific signals in e-commerce scenario. 
Specifically, we propose three different types of auxiliary tasks based on reconstruction, prediction, and contrastive learning respectively.
The reconstruction-based auxiliary task learns a list-wise representation that can be used to reconstruct key information in the original input. 
The prediction-based auxiliary task predicts the reward calculated based on user's feedback (e.g., click or pull-down). 
The contrastive-learning based auxiliary task aggregates  representations of similar state-action pairs and distinguishes representations of different state-action pairs.

We evaluate method using real-world dataset provided by Meituan food delivery platform. Both offline experimental results and online A/B test show that the proposed auxiliary tasks for ads allocation could effectively accelerate the list-wise representation learning of the agent and achieve significant improvement in terms of platform revenue. 

The contributions of this paper are summarized as follows:
\setlength{\parskip}{0.2cm plus4mm minus3mm}
\begin{itemize}[leftmargin=*]
  \item We introduce a unified RL framework to learn the list-wise representation for ads allocation. This solution enables us to handle list-wise representation learning efficiently in high-dimensional space.
  
  \item We design three novel auxiliary tasks, which effectively utilize the side information in ads allocation scenario to aid the learning of agent.
  
  \item We conduct extensive experiments on real-world dataset collected from Meituan food delivery platform. The results verify that our method achieves better performance.
\end{itemize}

\section{Related Works}
\subsection{Ads Allocation} 
\begin{figure}[tb]
  \centering
  \includegraphics[width=\linewidth]{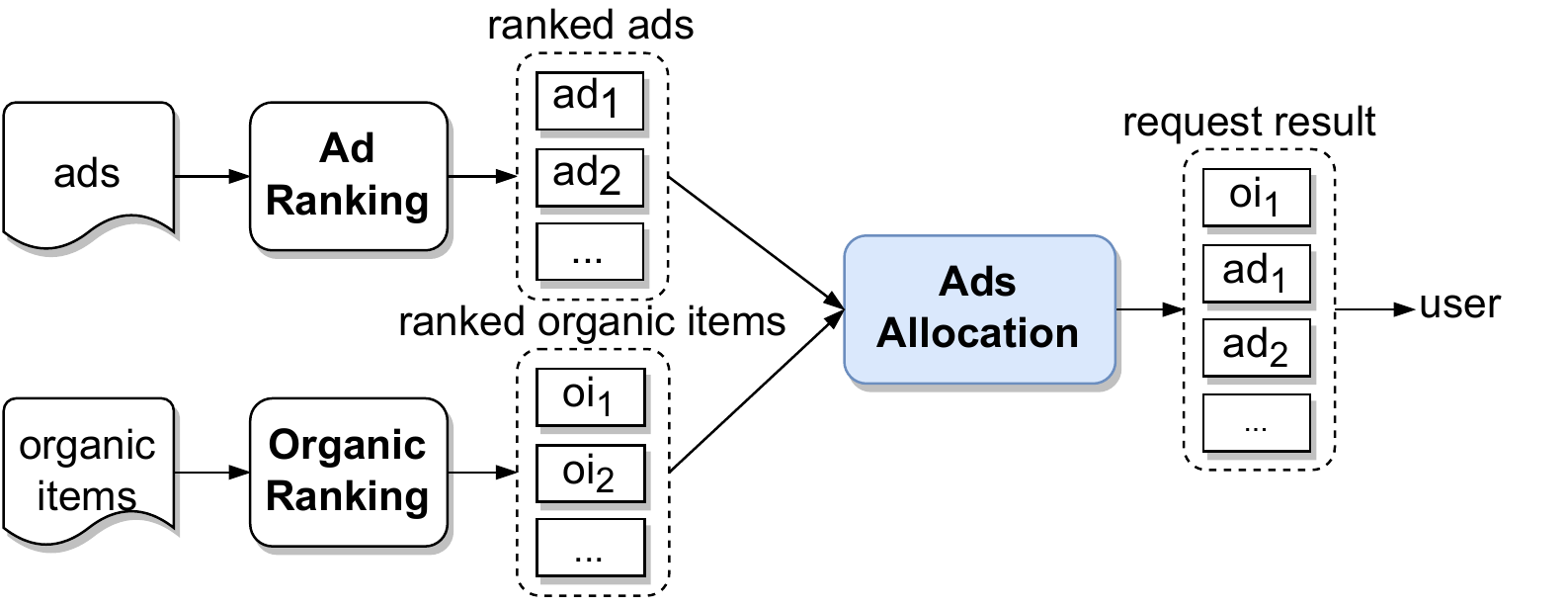}
  \Description{Structure of an ads allocation system. In ads allocation system, it takes the ranked ads and ranked organic items as input and outputs the allocated result.}
  \caption{Structure of an ads allocation system. In ads allocation system, it takes the ranked ads and ranked organic items as input and outputs the allocated result.}
  \label{fig:1}
\end{figure}

As shown in Figure \ref{fig:1}, the ads allocation system takes ranked ads list and ranked organic items list as input, and outputs a mixed list of the two \cite{liao2021cross,yan2020LinkedInGEA}. Traditional strategy for ads allocation is to display ads at fixed slots \cite{ouyang2020minet,li2020deep}. However, allocating ads to pre-determined slots may lead to suboptimal overall performance. Recently, dynamic ads allocation strategies \cite{yan2020LinkedInGEA,liao2021cross, zhao2020jointly,xie2021hierarchical}, which adjust the number and slots of ads according to the interest of users, have received growing attention. According to whether RL is used, existing dynamic ads allocation strategies can be roughly categorized into two categories \cite{liao2021cross}: non RL-based and RL-based.

Non RL-based dynamic ads allocation strategies usually use classical algorithms (e.g., Bellman-Ford algorithm, unified ranking score function and so on) to allocate ad slots. For instance, 
\citet{koutsopoulos2016optimal} define ads allocation as a shortest-path problem on a weighted directed acyclic graph where nodes represent ads or slots and edges represent expected revenue and use Bellman-Ford algorithm to find the shortest path as result.
Meanwhile, \citet{yan2020LinkedInGEA} propose a unified ranking score function which uses the interval between adjacent ads as an important factor.
But these methods cannot effectively utilize rich features hidden in items, resulting in poor scalability and iteration.

Since the feed continuously displays items to the user page by page, RL-based dynamic ads allocation strategies model the problem in feed as an MDP and solved it with different RL techniques. According to whether the input of the agent for ads allocation is an ad or an entire screen at a time, we divide these methods into point-wise allocation methods and list-wise allocation methods. 
For instance, \citet{xie2021hierarchical} propose a representative point-wise allocation method using hierarchical RL, where the low-level agent generates a personalized channel list (e.g., ad channel or organic item channel) and the high-level agent recommends specific items from heterogeneous channels under the channel constraints.
As for list-wise allocation methods, 
\citet{zhao2019deep} propose a DQN architecture to determine the optimal ads and ads position jointly. \citet{liao2021cross} propose a novel architecture named CrossDQN, which constructs the list-wise representation to model the arrangement signal hidden in the item arrangement.

Compared with point-wise allocating, list-wise allocation methods \cite{liao2021cross, liao2022deep} allocate multiple slots at a time,
which not only introduces the list-wise representation capabilities, 
but also induces several challenges including: larger state-action space, harder-to-learn representation and lower sampling efficiency.
To tackle these challenges, we focus on improving performance of RL-based dynamic ads allocation strategies by learning an efficient and generalizable list-wise representation. 
Although there has been efforts to apply representation learning in e-commerce scenarios such as CTR prediction \cite{ni2021effective, zhu2021representation,ouyang2019representation}, to the best of our knowledge, our approach is the first attempt to use representation learning in ads allocation. 

\subsection{Representation Learning} 
Auxiliary tasks are almost the benchmark for representation learning in RL nowadays \cite{rcrl,ha2018recurrent,guo2020bootstrap}. Specifically, the auxiliary task can be used for both the model-based setting and the model-free setting \cite{rcrl}. In the model-based settings, world models can be used as auxiliary tasks to achieve better performance \cite{franccois2019combined, hafner2019dream, hafner2019learning}. 
Since there are complex components e.g., the latent transition or reward module in the world model, these methods are empirically unstable to train and relies on different regularizations to converge \cite{rcrl}. In the model-free settings, various auxiliary tasks are constructed to improve performance \cite{shelhamer2016loss, guo2020bootstrap,mazoure2020deep,rcrl}. According to the different objectives of auxiliary tasks, existing auxiliary tasks in model-free RL frameworks can be roughly categorized into three types, reconstruction-based \cite{finn2016deep, ha2018recurrent}, prediction-based \cite{jaderberg2016reinforcement, van2018representation, guo2020bootstrap} and contrastive-learning based \cite{laskin2020curl,rcrl,anand2019unsupervised}.

As for reconstruction-based auxiliary tasks, the learning of representation is assisted by an encoder-decoder structure with the goal of minimizing the reconstruction error.
For instance, \citet{finn2016deep} present an auxiliary task for learning state representation using deep spatial autoencoders and \citet{ha2018recurrent} use variational autoencoder to accelerate the learning of representation layer. 
In many interesting environments reward is encountered very sparsely, making it difficult for the agent to learn.
So the prediction-based auxiliary tasks are proposed to remove the perceptual sparsity of rewards and rewarding states and used to aid the training of an agent. For example, \citet{jaderberg2016reinforcement} propose an auxiliary task to predict the onset of immediate reward given some historical context. \citet{van2018representation} propose a universal unsupervised learning approach to extract useful representation and hope the learned representation can predict the representation of the state of the subsequent steps respectively. Contrastive learning has seen dramatic progress recently, and been introduced to learn state representation. For instance, \citet{laskin2020curl} adopt a contrastive auxiliary task to accelerate representation learning by leveraging the image augmentation. \citet{rcrl} leverage return to construct a contrastive auxiliary task for speeding up the main RL task.

In this paper, we propose three different types of auxiliary tasks for ads allocation based on prior domain knowledge and combine them in a single framework, which can be easily applied to the existing RL-based ads allocation strategies and improve the performance.

\section{Problem Formulation}
\label{sec:problem}
We adopt a great paradigm for problem formulation, also used in recent related works such as CrossDQN \cite{liao2021cross}.  
When a user browses in feed, the platform displays items to the user page by page in one request. Each page contains $K$ slots and the task is to allocate ad slots for each page in the request sequentially.
Mathematically, the ads allocation problem is formulated as an MDP, which can be denoted by a tuple ($\mathcal{S}$, $\mathcal{A}$, $r$, ${P}$, $\gamma$), as follows:

\begin{itemize}[leftmargin=*]
  \item \textbf{State space $\mathcal{S}$}.
  A state $s\in\mathcal{S}$ consists of the information of candidate items, the user and the context. 
  The candidate items consist of the ads list and the organic items list which are available in current page $t$. 
  Each item has $H_i$ sparse features (e.g., whether it is free of delivery fee, whether it is in promotion, whether it is a brand merchant and so on).
  Each user has $H_u$ sparse features (e.g., age, gender and so on) and $N_b$ historical behaviors. 
  The context contains $H_c$ sparse features (e.g., order time, order location and so on).
  With embedding, these large-scale sparse features are transformed into low-dimensional dense vectors.
  See more details in Section \ref{sec:base_agent}.
  \item \textbf{Action space $\mathcal{A}$}. An action $a \in \mathcal{A}$ is the decision whether to display an ad on each slot in current page, which is formulated as follows:
  \begin{equation}
    \begin{aligned}
      a=(x_{1}, x_{2}, \ldots, x_{K}),\ \ \forall x_k \in \{0,1\},
    \end{aligned}
  \end{equation}
  where $x_{k}=1$ means to display an ad on the $k$-th slot and $x_{k}=0$ means to display an organic item on the $k$-th slot. 
  In our scenario, we do not change the sequence of the items within ads list and organic items list when allocating slots. 
  \item \textbf{Reward $r$}. After the system takes an action in one state and generates a page, a user browses this page of the mixed list and gives a feedback. The reward includes platform revenue and user experience, as follows:
  \begin{equation}
    \begin{aligned}
  r(s,a) &= r^\text{ad}+r^\text{fee}+\eta r^\text{ex}
    \end{aligned}
    \label{eq:eta}
  \end{equation}
  
  where $r^\text{ad},r^\text{fee} $ and $r^\text{ex}$ denote the ads revenue, service fee and user experience score of this page respectively. $r^\text{ex}$ is set to $2,1,0$ when the user places an order, clicks, leaves, respectively. $\eta$ is the coefficient used to balance platform revenue and user experience.

  \item \textbf{Transition probability ${P}$}.
  $P(s_{t+1}|s_t,a_t)$  defines the state transition probability from $s_t$ to $s_{t+1}$ after taking the action $a_t$, where $t$ is the index for the page in a request.
  When the user pulls down to the first item in the next page, the state $s_{t}$ transits to the state of next page $s_{t+1}$.
  Since seeing the same items in the same request causes awful user experience, the items selected by $a_t$ are removed from the next state $s_{t+1}$.
  When the user no longer pulls down, the transition terminates. 
  
  \item \textbf{Discount factor $\gamma$}. The discount factor $\gamma \in [0, 1]$ balances the short-term and long-term rewards.
\end{itemize}

Given the MDP formulated as above, the objective is to find an ads allocation policy $\pi: \mathcal{S} \rightarrow \mathcal{A}$ to maximize the total reward.
In this paper, we mainly focus on how to design auxiliary tasks to accelerate representation learning and improve the performance.

\begin{figure*}[tb]
  \centering
  \includegraphics[width=1\linewidth]{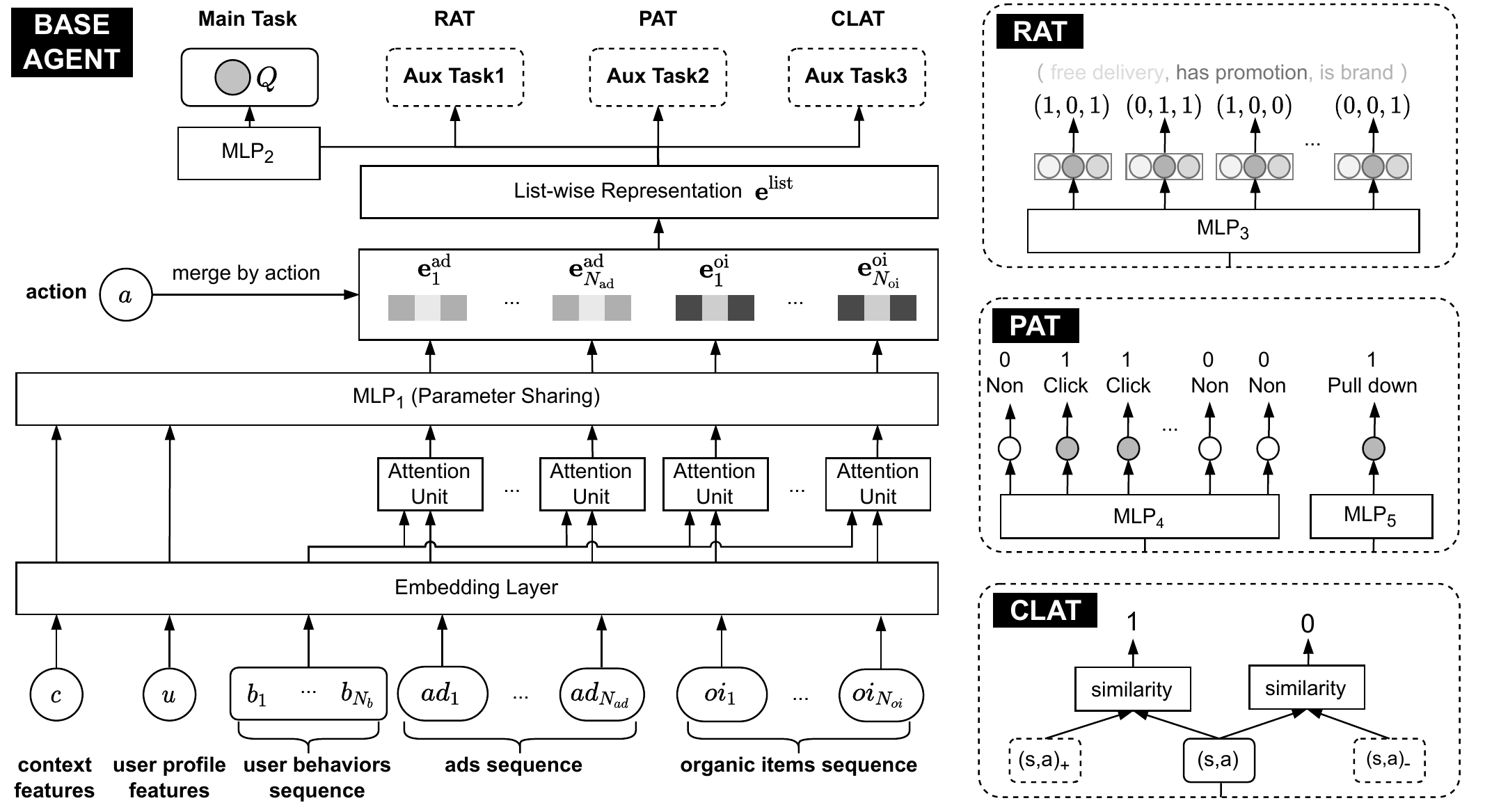}
  \caption{
  Our method consists of a base agent and three different types of auxiliary tasks. The base agent takes a state information and an action as input and outputs the Q-value. Three different types of auxiliary tasks are designed to accelerate the base agent training. Best view in color.
  }
  \label{fig:fig1}
\end{figure*}

\section{Methodology}
As shown in Figure \ref{fig:fig1}, our method consists of a base agent and three different types of auxiliary tasks. The base agent first takes a state and an action as input to generate the list-wise representation and uses the representation to predict the corresponding Q-value.
Specifically, these three different auxiliary tasks are designed to accelerate the learning of list-wise representation: 
i) The Reconstruction-based Auxiliary Task (RAT) adopts the encoder-decoder structure to reconstruct key information in the list-wise representation. ii) The Prediction-based Auxiliary Task (PAT) utilizes user behaviors to alleviate the sparse reward problem and guide the learning of the list-wise representation. iii) The Contrastive-Learning based Auxiliary Task (CLAT) is a method that constructs positive sample pairs for ads allocation and optimizes a contrastive loss to help the learning of list-wise representation.
We describe the above components in detail in the following subsections. 

\subsection{Base Agent}
\label{sec:base_agent}
The base agent takes a state and an action as input to generate a list-wise representation and outputs the corresponding Q-value as depicted in.
Notice that, the base agent is mainly used to illustrate the general practice of RL-based ads allocation rather than the main focus of this paper.  Therefore, some complex but effective modules (e.g., Dueling DQN \cite{wang2016duelingDQN,liao2021cross}, multi-channel attention unit \cite{liao2021cross}, multi-channel interaction module \cite{liao2022deep}) are not mentioned. But they can be easily added to the base agent to further improve the overall performance in experiment.
See more details in Section \ref{sec:exp_set}.

Here we first employ an embedding layer to extract the embeddings from raw inputs. 
Mathematically, sparse features can be represented by $\mathbf{E} \in \mathbb{R}^{H \times d_{e}}$, where $H$ is the number of sparse features (i.e., $H_i, H_u, H_c$ for item, user and context, respectively) and $d_{e}$ is the dimension of embedding. Then we flatten each matrix $\mathbf{E}$ and represent it as $\mathbf{e}$. 
We denote the embeddings for ads, organic items, historical behaviors of the user, the user profile, the context as $\{\mathbf{e}_i^\text{ad}\}_{i=1}^{N_\text{ad}}$, $\{\mathbf{e}_i^\text{oi}\}_{i=1}^{N_\text{oi}}$, $\{\mathbf{e}_i^\text{b}\}_{i=1}^{N_\text{b}}$, $\mathbf{e}^\text{u}$, and $\mathbf{e}^\text{c}$ respectively, where the subscript $i$ denotes the index within the list and $N_\text{ad}$, $N_\text{oi}$, and $N_\text{b}$ are the number of ads, organic items, and historical behaviors.

After embedding, we adopt a target attention network \cite{vaswani2017attention,zhou2018DIN} followed by a Multi-Layer Perception (MLP) to generate the representation of each item:
\begin{equation}
  \begin{aligned}
  \mathbf{e}_j^\text{ad}& \leftarrow \text{MLP}_1\bigg( \text{Att} \Big( \mathbf{e}^\text{ad}_{j}, \{\mathbf{e}^\text{b}_{i} \}_{i=1}^{N_\text{b}} \Big) ||  \mathbf{e}^\text{ad}_{j} ||\mathbf{e}^\text{u}|| \mathbf{e}^\text{c} \bigg), \forall j\in [N_\text{ad}]; \\
  \mathbf{e}_j^\text{oi}& \leftarrow \text{MLP}_1\bigg( \text{Att} \Big( \mathbf{e}^\text{oi}_{j}, \{\mathbf{e}^\text{b}_{i} \}_{i=1}^{N_\text{b}} \Big)|| \mathbf{e}^\text{oi}_{j} ||\mathbf{e}^\text{u}|| \mathbf{e}^\text{c} \bigg), \forall j\in [N_\text{oi}],
\end{aligned}
\end{equation}
where $||$ denotes concatenation, $\text{Att} \Big( \mathbf{e}^\text{ad}_{j}, \{\mathbf{e}^\text{b}_{i} \}_{i=1}^{N_\text{b}} \Big)$ is the target attention unit which calculates the attention weight of each historical behavior and generates a weighted behavior representation. The embeddings of item, user profile and the context are concatenated with the behavior representation generated from target attention unit to generate the representation of each item (i.e., each ad and each organic item).

Afterwards, the representations of ads and organic items are selected and concatenated according to the action. For example, when action $a=(0,0,1,0,1,0,0,0,0,1)$\footnote{$K$ in this example is $10$.}, the first three of ranked ads list and the first seven of ranked organic items list are selected and concatenated as follows:
\begin{equation}
  \begin{aligned}
\mathbf{e}^{\text{list}} = \mathbf{e}_1^\text{oi} ||\mathbf{e}_2^\text{oi} || \mathbf{e}_1^\text{ad} || \mathbf{e}_3^\text{oi} || \mathbf{e}_2^\text{ad} ||\mathbf{e}_4^\text{oi} ||\mathbf{e}_5^\text{oi} ||\mathbf{e}_6^\text{oi} ||\mathbf{e}_7^\text{oi} || \mathbf{e}_3^\text{ad},
  \end{aligned}
\end{equation}

 Finally, the base agent feeds the list-wise representation into an MLP  and outputs the Q-value:
\begin{equation}
  \begin{aligned}
Q(s,a) = \text{MLP}_{\text{2}}\big(\mathbf{e}^{\text{list}}\big).
  \end{aligned}
\end{equation}

\subsection{Reconstruction-based Auxiliary Task}
Reconstruction can prevent key information from being lost in the representation \cite{mirowski2016learning,van2020mdp,laskin2020curl,hessel2019multi}. 
In Meituan food delivery platform, users care about some aspects when they browses and place orders,
e.g., the delivery fee, promotion, brand and so on.
The above aspects greatly influence the user experience thus are of high correlation with the behaviors of users.

There is a strong correlation between these key information and behavior of users. Preventing them from being lost in the list-wise representation can effectively improve the performance of the agent. To this end, we select the top $M$ most concerned factors as labels\footnote{The ranking score $\beta$ of different factors is determined based on the prior survey on user preferences.} to build the reconstruction-based auxiliary task. 

Specifically, RAT takes the list-wise representation as input and $M$ types of binary features as labels. The decoder network outputs the predicted values for each slot, as follows:
\begin{equation}
  \begin{aligned}
\hat y_{k;m} =  \text{MLP}_3\big(\mathbf{e}^{\text{list}}\big), \ \ \forall k \in [K], \forall m \in [M],
  \end{aligned}
\end{equation}
where $\hat y_{k;m} $ is the $m$-th predicted value for the item on the $k$-th slot. 

The reconstruction-based auxiliary loss is:
\begin{equation}
\begin{aligned}
L_{\text{RAT}} =  \sum_{m=1}^{M} \beta_{m} \cdot \Big( \sum_{k =1}^K \text{CE}(y_{k;m},\hat y_{k;m}) \Big).
\end{aligned}
\end{equation}
where $\beta_{m}$ is the weight of the $m$-th key information, $y_{k;m}$ is the label for the $m$-th key information of the item on the $k$-th slot, and the cross entropy $\text{CE}(y,\hat y)$ is defined as:
\begin{equation}
\begin{aligned}
  \text{CE}(y,\hat y) = -y\log ( \hat y) - (1-y)\log(1-\hat y)
\end{aligned}
\end{equation}

\subsection{Prediction-based Auxiliary Task}
  In practice, the base agent may suffer from sample inefficiency due to the natural sparsity of reward compared to the tremendous search space. 
  Therefore, we incorporate supervised signals based on user behaviors to jointly guide the agent in training. 
  There are two main types of user behaviors: click and pull-down. The former can be used to predict the reward of the current request and the latter determines whether the trajectory terminates.

  Specifically, the click-based prediction task takes the list-wise representation as input and outputs the predicted click-through rates (CTR) for each slot:
  \begin{equation}
  \begin{aligned}
\hat z_k =  \text{MLP}_4\big(\mathbf{e}^{\text{list}}\big), \ \ \forall k \in [K].
  \end{aligned}
\end{equation}

In our scenario, since the transition of adjacent requests is reflected in the probability of pull-down, designing an auxiliary task to predict whether there is an pull-down or not can help the list-wise representation to embed the impact of the current request on subsequent requests. Analogously, the pull-down based prediction task takes the list-wise representation as input and outputs the probability of the user's pull-down as follows:
\begin{equation}
  \begin{aligned}
    \hat p = \text{MLP}_5\big(\mathbf{e}^{\text{list}}\big).
  \end{aligned}
\end{equation}

The prediction-based auxiliary loss is:
\begin{equation}
  \begin{aligned}
L_{\text{PAT}} = \sum_{k=1}^K\text{CE}(z_k, \hat z_k) + \text{CE}(p, \hat p)
  \end{aligned}
\end{equation}
where $z_k$ indicates the item in the $k$-th slot is clicked or not by the user $u$ and $p$ indicates whether there is a pull-down or not. The prediction-based auxiliary loss directly optimizes the list-wise representation through the supervision information based on user behaviors, which can make training more robust and accelerate model convergence in training.

 \begin{figure}[b]
  \centering
  \includegraphics[width=\linewidth]{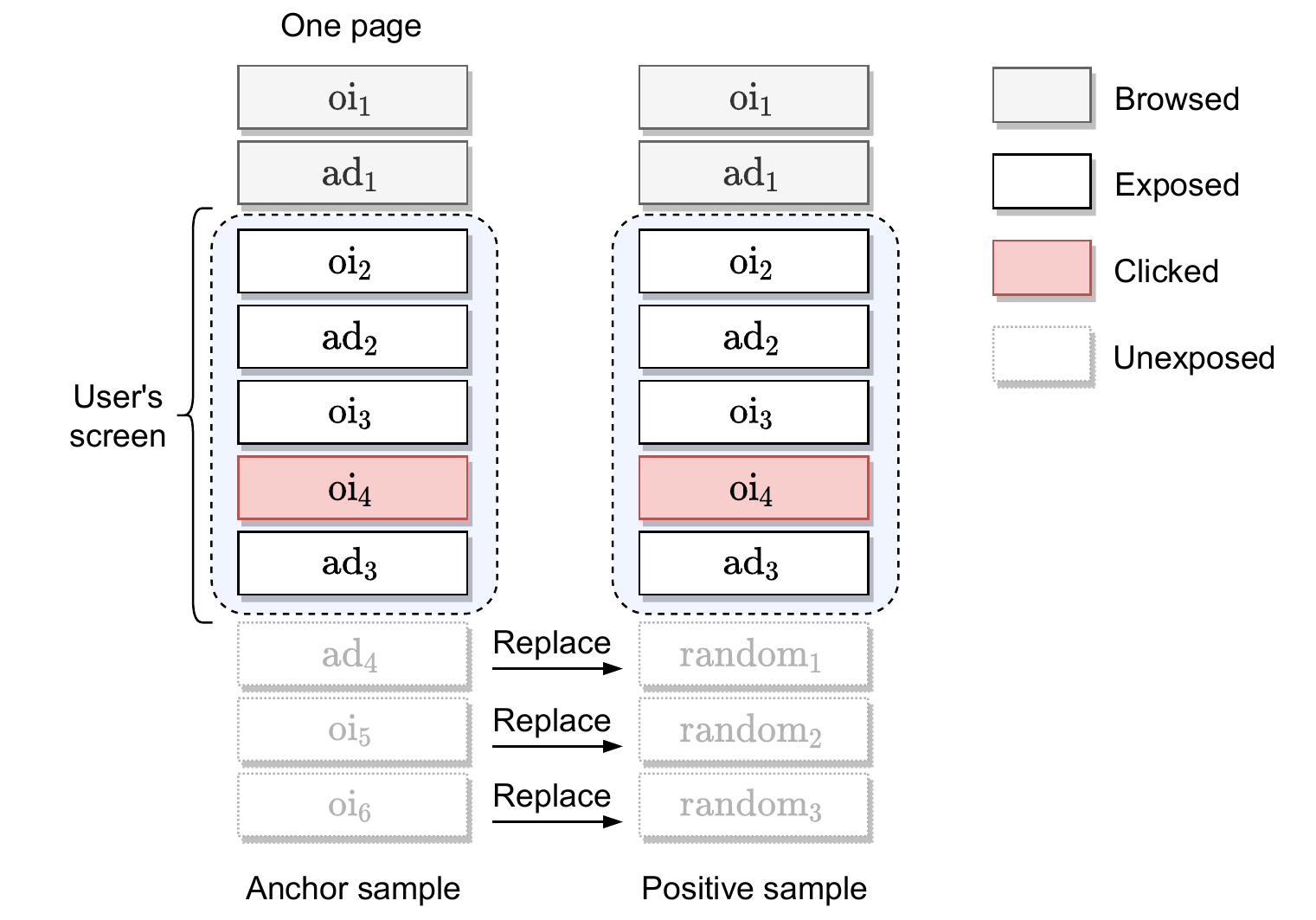}
  \caption{An instance of constructing positive sample for contrastive learning.}
  \label{fig:3}
\end{figure}

\subsection{Contrastive-Learning based Auxiliary Task}
The main idea of most contrastive-learning based auxiliary tasks is to hope that the representation of anchor sample is closer to the representation of positive samples and farther from the representation of negative samples in the latent vector space \cite{rcrl,anand2019unsupervised,sermanet2018time,dwibedilearning,aytar2018playing}.
Here we introduce a contrastive learning based auxiliary task to improve the differentiation of the representation between different types of state-action pairs. Taking one sample as an anchor sample, we define a sample with similar representation in state-action space as a positive sample and a sample with different representation in state-action space as a negative sample. The following is the detail of constructing positive and negative samples.

Firstly, we construct positive sample based on user behavior. For instance, as shown in Figure \ref{fig:3}, if the user scrolls to the 7th slot in current page and gives a feedback, it reflects that the items other than the first seven have little influence on user since the user may have not seen them when making the decision. Therefore, we replace the rest of the items in current page to generate a positive sample whose representation in the state-action space should be close to the anchor sample. For negative samples, we randomly sample from other requests. Finally, the contrastive-learning based auxiliary loss \cite{sohn2016improved} is calculated as:
\begin{equation}
  \begin{aligned}
L_{\text{CLAT}} = - \log \Big(\frac{\exp \big(s ({\mathbf{e}^{\text{list}},\mathbf{e}^{\text{list}}_{+}}) \big)}{ \exp \big( s ({\mathbf{e}^{\text{list}},\mathbf{e}^{\text{list}}_{+}}) \big)+ \sum_{j=1}^{L-1}  \exp \big(s(\mathbf{e}^{\text{list}},\mathbf{e}^{\text{list}}_j)\big)}\Big),
  \end{aligned}
\end{equation}
where $\mathbf{e}^{\text{list}}_{+}$ is the representation of positive sample, $L-1$ is the number of negative samples, and the consine similarity of two representations is calculated as follows:
\begin{equation}
  \begin{aligned}
    s({\mathbf{e}^{\text{list}}_i,\mathbf{e}_j^{\text{list}}})=\frac {\mathbf{e}^{\text{list}}_i \cdot \mathbf{e}^{\text{list}}_j} {|\mathbf{e}_i^{\text{list}}| \cdot |\mathbf{e}^{\text{list}}_j|}.
  \end{aligned}
\end{equation}

\subsection{Offline Training}
\begin{algorithm}[tbp]  
  \renewcommand\arraystretch{1.1}
  \caption{Offline training of our method}
  \label{alg:offline}
  \begin{algorithmic}[1] 
    \State Offline data $D=\{(s,a,r,s')\}$ 
    \State Extract labels for reconstruction and prediction
    \State Construct positive example and sample negative samples set for each sample $(s,a,r,s')$
    \State Initialize a value function $Q$ with random weights
    \Repeat
        \State Sample a batch $B$ of $(s,a,r,s')$ from $D$
        \State Update network parameters by minimizing $L(B)$ in \eqref{eq:loss}
    \Until Convergence
  \end{algorithmic}  
  \end{algorithm}

\begin{table*}[tb]

  \caption{The performance of different models. The results
   are presented in the form of mean ($\pm$ standard deviation). The improvement indicates the improvement of our method over the best baselines.}
  \label{result}
  \renewcommand\arraystretch{1.15}
  \setlength{\tabcolsep}{4.9mm}{
  \begin{tabular*}{0.92\textwidth}{l|cc|cc}
  \hline

  \multirow{2}{*}{Method}  & \multicolumn{2}{c|}{Revenue Indicators}                       & \multicolumn{2}{c}{Experience Indicators}         \\
                     & $R^\text{ad}$               & $R^\text{fee}$              & $R^\text{cxr}$              & $R^\text{ex}$               \\
  \hline \hline
  HRL-Rec               & 0.2390\ ($\pm$0.0035) & 0.2662\ ($\pm$0.0053) & 0.2441\ ($\pm$0.0013) & 0.9530\ ($\pm$0.0038) \\
 DEAR               & 0.2402\ ($\pm$0.0029) & 0.2679\ ($\pm$0.0067) & 0.2487\ ($\pm$0.0007) & 0.9621\ ($\pm$0.0024) \\
 CrossDQN               & 0.2465\ ($\pm$0.0006) & 0.2740\ ($\pm$0.0009) & 0.2508\ ($\pm$0.0008) & 0.9667\ ($\pm$0.0036) \\
  \hline   \hline
  DSAE        & 0.2466\ ($\pm$0.0018) & 0.2741\ ($\pm$0.0052) & 0.2508\ ($\pm$0.0011) & 0.9635\ ($\pm$0.0047) \\
  UNREAL        & 0.2471\ ($\pm$0.0012) & 0.2744\ ($\pm$0.0038) & 0.2511\ ($\pm$0.0006) & 0.9645\ ($\pm$0.0022) \\
  RCRL        & 0.2482\ ($\pm$0.0007) & 0.2765\ ($\pm$0.0022) & 0.2521\ ($\pm$0.0006) & 0.9648\ ($\pm$0.0023) \\
  \hline   \hline
  \textbf{Our Method}       & \textbf{0.2541\ ($\pm$0.0006)} & \textbf{0.2815\ ($\pm$0.0024)} & \textbf{0.2554\ ($\pm$0.0008)} & \textbf{0.9718\ ($\pm$0.0031)}   \\
 \ \ \ - w/o $\text{RAT}$    & 0.2491\ ($\pm$0.0003) & 0.2761\ ($\pm$0.0009) & 0.2538\ ($\pm$0.0004) & 0.9701\ ($\pm$0.0015) \\
 \ \ \ - w/o $\text{PAT}$    & 0.2487\ ($\pm$0.0011) & 0.2756\ ($\pm$0.0028) & 0.2535\ ($\pm$0.0009) & 0.9694\ ($\pm$0.0042) \\
 \ \ \ - w/o $\text{CLAT}$    & 0.2483\ ($\pm$0.0010) & 0.2768\ ($\pm$0.0013) & 0.2531\ ($\pm$0.0010) & 0.9688\ ($\pm$0.0044) \\
 \ \ \ - w/o $\text{all AT}$   & 0.2464\ ($\pm$0.0051) & 0.2739\ ($\pm$0.0007) & 0.2506\ ($\pm$0.0008) & 0.9665\ ($\pm$0.0036) \\
  \hline \hline
  Improvement        & \textbf{2.38\%}      & \textbf{1.81\%}      & \textbf{1.31\%}      & \textbf{0.73\%}      \\
  \hline
  \end{tabular*}}
  \label{tab:all_res}
\end{table*}

\noindent We follows the offline RL paradigm, and the process of offline training is shown in Algorithm \ref{alg:offline}.
We train our agent based on an offline dataset $D$ generated by an online random exploratory policy $\pi_b$. 
For each iteration, we sample a batch of transitions $B$ from the offline dataset and update the agent using gradient back-propagation w.r.t. the loss:
\begin{equation}
\label{eq:loss}
L(B) = \frac{1}{|B|}\sum_{(s,a,r,s')\in B} \Big( L_\text{DQN} + \alpha_1 \cdot L_{\text{RAT}} + \alpha_2 \cdot L_{\text{PAT}} + \alpha_3 \cdot L_{\text{CLAT}} \Big),
\end{equation}
where $L_\text{DQN}$ is the same loss function as the loss in DQN \cite{mnih2015human}, and $\alpha_1,\alpha_2,\alpha_3$ are the coefficients to balance the four losses.
Specifically, 
\begin{equation}
L_\text{DQN} =  \Big( r + \gamma \max_{a'\in\mathcal{A}} Q(s', a') - Q(s,a) \Big) ^ 2.
\end{equation}

\subsection{Online Serving} 
\label{sec:online imp} 
\noindent We illustrate the process of online serving in Algorithm \ref{alg:online1}. 
  In the online serving system, the agent selects the action with the highest reward 
  based on current state and converts the action to ad slots set for the output. When the user pulls down, the agent receives the state for the next page, and then makes a decision based on the information of next state.

  \begin{algorithm}[tb]  
    \renewcommand\arraystretch{1.1}
  \renewcommand{\algorithmicrequire}{\textbf{Input:}}
  \renewcommand{\algorithmicensure}{\textbf{Output:}}
      \caption{Online inference of our method}
      \label{alg:online1}
      \begin{algorithmic}[1] 
        \State Initial state $s_0$
        \Repeat
        \State Generate $a^*_{t} = \arg \max_{a\in \mathcal{A}} Q(s_t,a)$
        \State Allocate ad slots following  $a^*_{t}$
        \State User pulls down 
        \State Observe the next state $s_{t+1}$
         \Until User leaves
      \end{algorithmic}  
    \end{algorithm} 

\section{Experiments}
In this section, we evaluate the proposed method\footnote{The code and data example are publicly accessible at \url{https://github.com/princewen/listwise_representation}} on real-world dataset,  with the aim of answering the following two questions:
i) How does our method perform compared with the baselines? 
ii) How do different auxiliary tasks and hyperparameter settings affect the performance of our method?

\subsection{Experimental Settings}
\label{sec:exp_set}
\subsubsection{Dataset}
Since there are no public datasets for ads allocation problem, we collect a real-world dataset by running a random exploratory policy on the Meituan platform during March 2021. As presented in Table \ref{dataset}, the dataset contains $12,729,509$ requests, $2,000,420$ users, $385,383$ ads and $726,587$ organic items. 
Notice that each request contains several transitions.

\subsubsection{Evaluation Metrics}

Since both user experience and platform revenue are important for ads allocation, similar to \cite{liao2021cross}, we evaluate the performance of methods with both revenue indicators and experience indicators. 
As for revenue indicators, we use ads revenue and service fee in a period to measure platform revenue, which is calculated as $R^\text{ad}= {\sum r^\text{ad}}/{N_\text{request}}$ and $R^\text{fee}= {\sum r^\text{fee}}/{N_\text{request}}$ (${N_\text{request}}$ is the number of requests).
As for experience indicators, we use the the global ratio of the number of orders to the number of requests and the  average user experience score (defined in Section \ref{sec:problem}) to measure the degree of satisfaction of the user demand, which is calculated as  $R^\text{cxr}={ N_\text{order} }/{ N_\text{request} }$ and $R^\text{ex}={\sum r^{\text{ex}} }/{ N_\text{request}}$ ($N_\text{order}$ is the number of orders).

\subsubsection{Hyperparameters}
We implement our method with TensorFlow and apply a gird search for the hyperparameters. $\eta$ is $0.05$\footnote{Follow the experiment result in \cite{liao2021cross}.}, $\alpha_1$ is $0.01$, $\alpha_2$ is $0.05$, $\alpha_3$ is $0.05$, $K$ is $10$, $L$ is $10$, $M$ is $3$, the hidden layer sizes of all MLPs are $(128, 64, 32)$,
the learning rate is $10^{-3}$, the optimizer is Adam \cite{kingma2014adam} and the batch size is 8,192.

\subsection{Offline Experiment}
In this section, we train our method with offline data and evaluate the performance using an offline estimator. We use Cross DQN \cite{liao2021cross} as the base agent in this subsection to achieve further improvement. 
Through extended engineering, the offline estimator models the user preference and aligns well with the online service.

\begin{table}[tbp]
  \Description{Statistics of the dataset.}
  \caption{Statistics of the dataset.}
  \renewcommand\arraystretch{1.3}
  \centering
  \begin{tabular}{ccccc}
    \hline
  \#requests    & \#users  & \#ads   & \#organic items \\
  \hline
  12,729,509 & 2,000,420 & 385,383 & 726,587 \\
  \hline
  \end{tabular}
  \label{dataset}
\end{table}

\subsubsection{Baselines}
We compare our method with the following three representative RL-based dynamic ads allocation methods and three representative RL representation learning methods using different types of auxiliary tasks:
\begin{itemize}[leftmargin=*]
  \item \textbf{HRL-Rec} \cite{xie2021hierarchical}. 
  HRL-Rec is a typical RL-based ads allocation method, which divides the integrated recommendation into two levels of tasks and solves using hierarchical reinforcement learning. Specifically, the model first decides the channel (i.e., select an organic item or an ad) and then determine the specific item for each slot.

  \item \textbf{DEAR} \cite{zhao2019deep}. 
  DEAR is an advanced RL-based ads allocation method, which designs a deep Q-network architecture to determine three related tasks jointly, i.e., i) whether to insert an ad to the recommendation list, and if yes, ii) the optimal ad and iii) the optimal location to insert.

  \item \textbf{Cross DQN} \cite{liao2021cross}. Cross DQN is a state-of-the-art RL-based ads allocation method, which takes the crossed state-action pairs as input and allocates slots in one screen at a time. It designs some units (e.g., MCAU) to optimize the combinatorial impact of the items on user behavior.
  
  \item \textbf{DSAE} \cite{finn2016deep}. DSAE presents a reconstruction-based auxiliary task for representation learning in RL, which uses deep spatial autoencoders to learn the spatial feature representation. Here we take Cross DQN as agent and use the deep spatial autoencoders to build an auxiliary task, with the aim of helping the learning of list-wise representation.
  
  \item \textbf{UNREAL} \cite{jaderberg2016reinforcement}. UNREAL contains a prediction-based auxiliary task for representation learning in RL, which predicts the onset of immediate reward with some historical context. Here we take Cross DQN as agent and use the prediction-based auxiliary task in UNREAL to aid the training of an agent.
  
  \item \textbf{RCRL} \cite{rcrl}. RCRL proposes a return-based contrastive representation learning method for RL, which leverages return to construct a contrastive auxiliary task for speeding up the main RL task.  Here we also take Cross DQN as agent for equality and use the return-based contrastive loss to accelerate representation learning.
\end{itemize}

\begin{figure*}[tb] 
  \centering 
  \subfigure[{The reward curve of $\alpha_1$.}]{ 
    \centering
    \begin{minipage}{0.304\linewidth}
    \includegraphics[width=\textwidth]{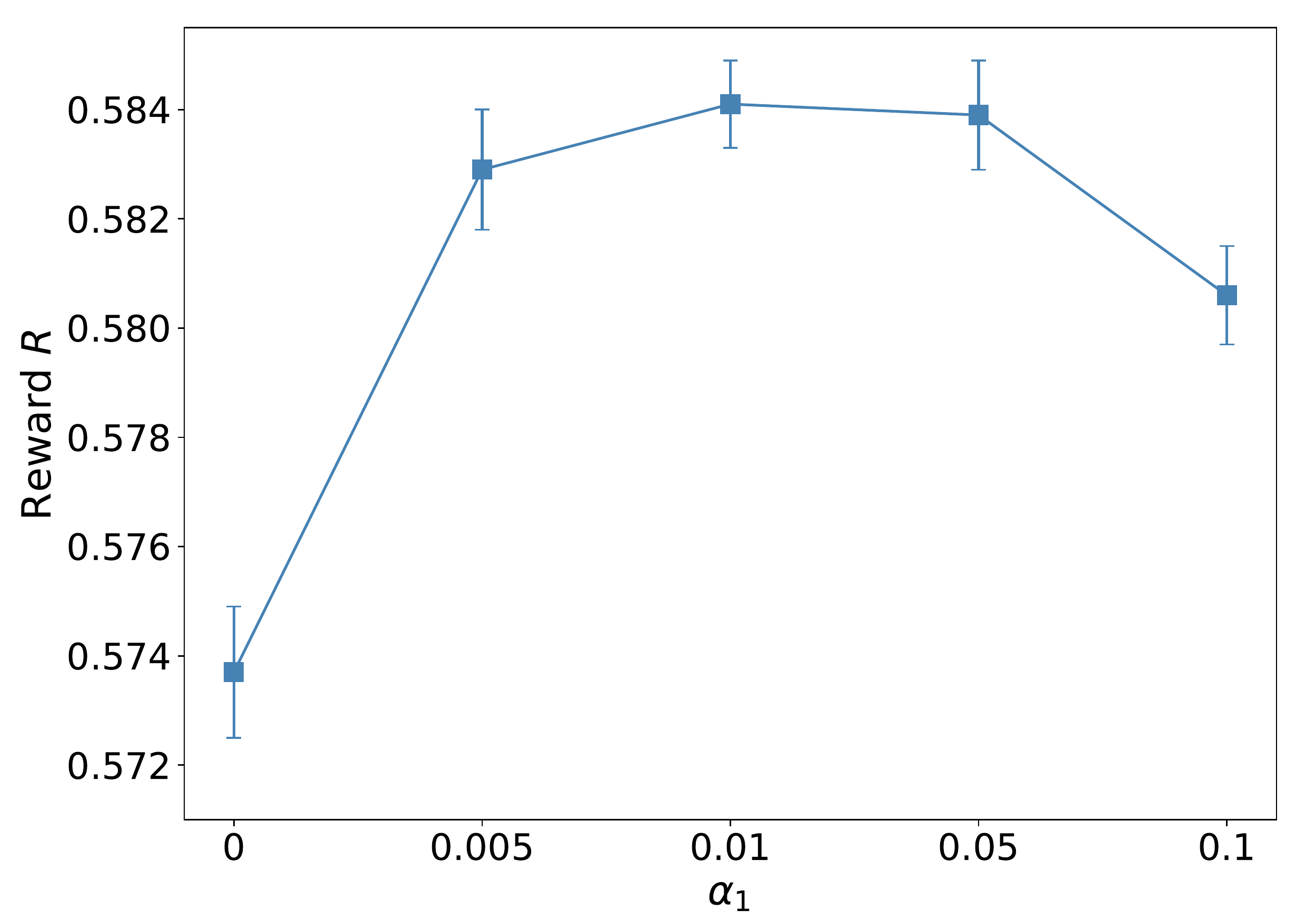} 
    \label{fig:subfig:a1} 
  \end{minipage}%
  } 
  \subfigure[{The reward curve of $\alpha_2$.}]{ 
    \centering
    \begin{minipage}{0.304\linewidth}
    \includegraphics[width=\textwidth]{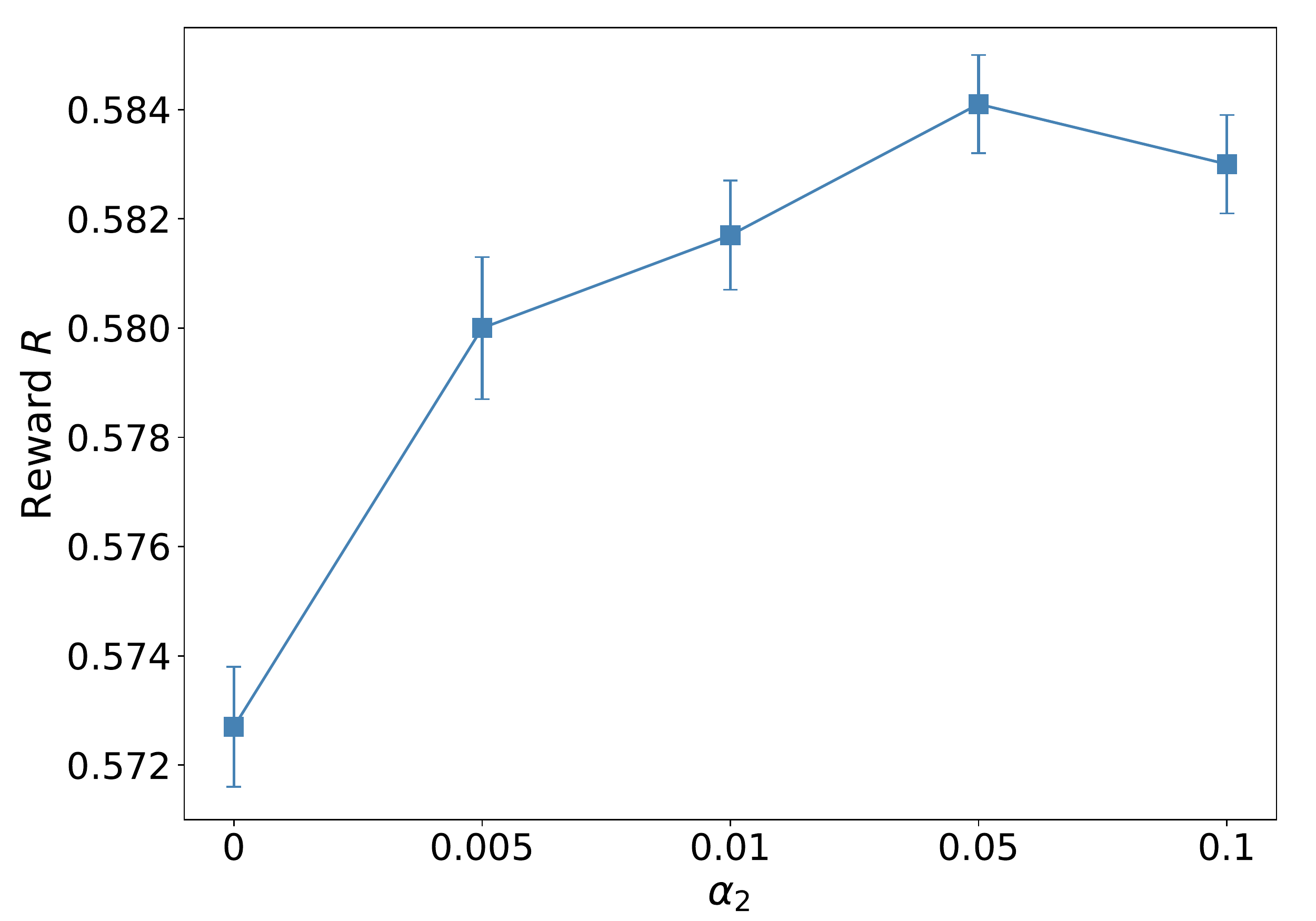} 
    \label{fig:subfig:a2} 
  \end{minipage}%
  } 
  \subfigure[{The reward curve of $\alpha_3$.}]{ 
    \centering
    \begin{minipage}{0.304\linewidth}
    \includegraphics[width=\textwidth]{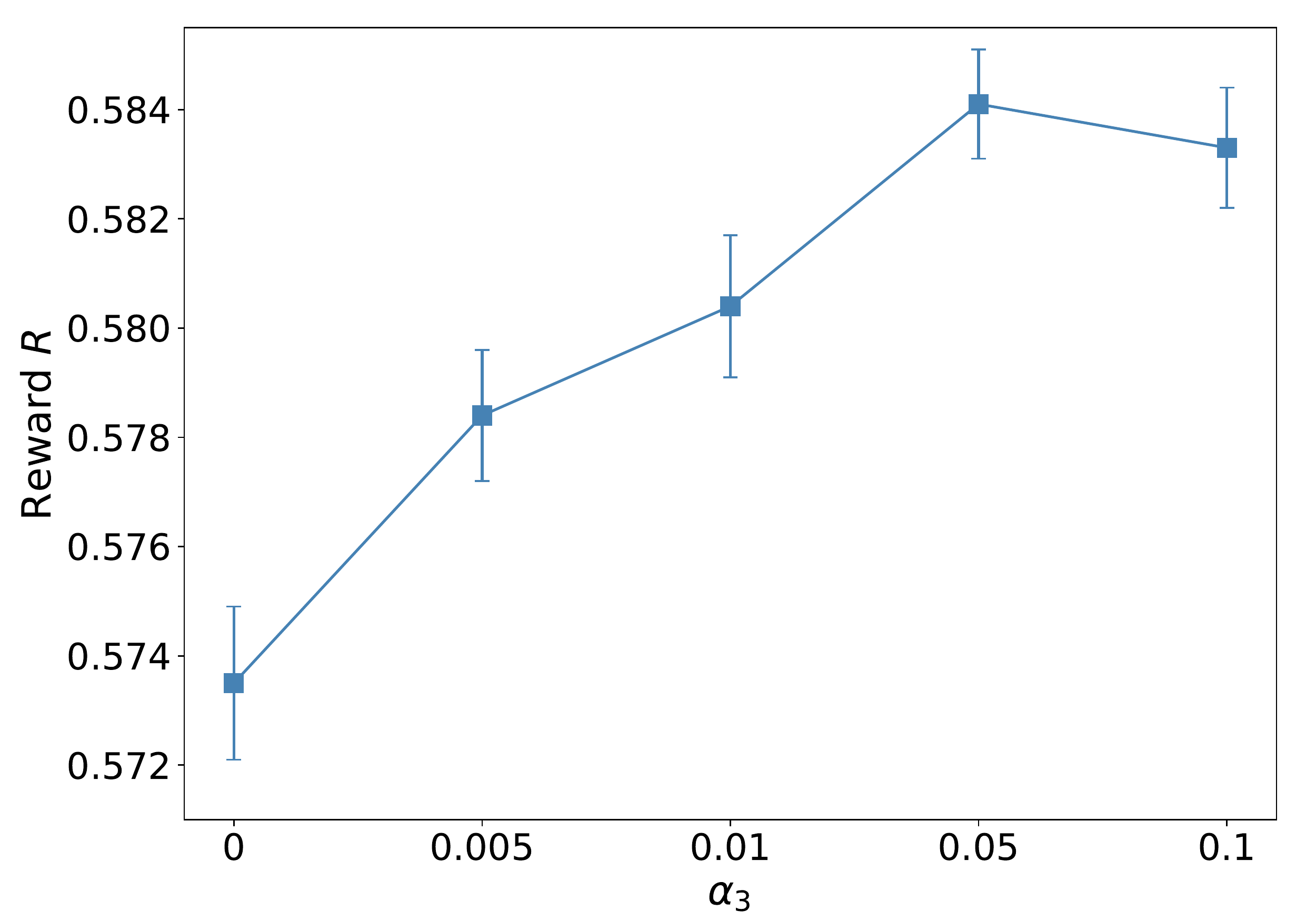}
    \label{fig:subfig:a3} 
  \end{minipage}%
  } 
  \Description{The experimental results on the sensitivity of $\alpha_1$, $\alpha_2$ and $\alpha_3$.}
  \caption{{The experimental results on the sensitivity of $\alpha_1$, $\alpha_2$ and $\alpha_3$.}
  }
  \label{fig:alpha}
\end{figure*}

\begin{figure*}[tb] 
  \centering 

  \subfigure[{The reward curve of $M$.}]{ 
    \centering
    \begin{minipage}{0.303\linewidth}
    \includegraphics[width=\textwidth]{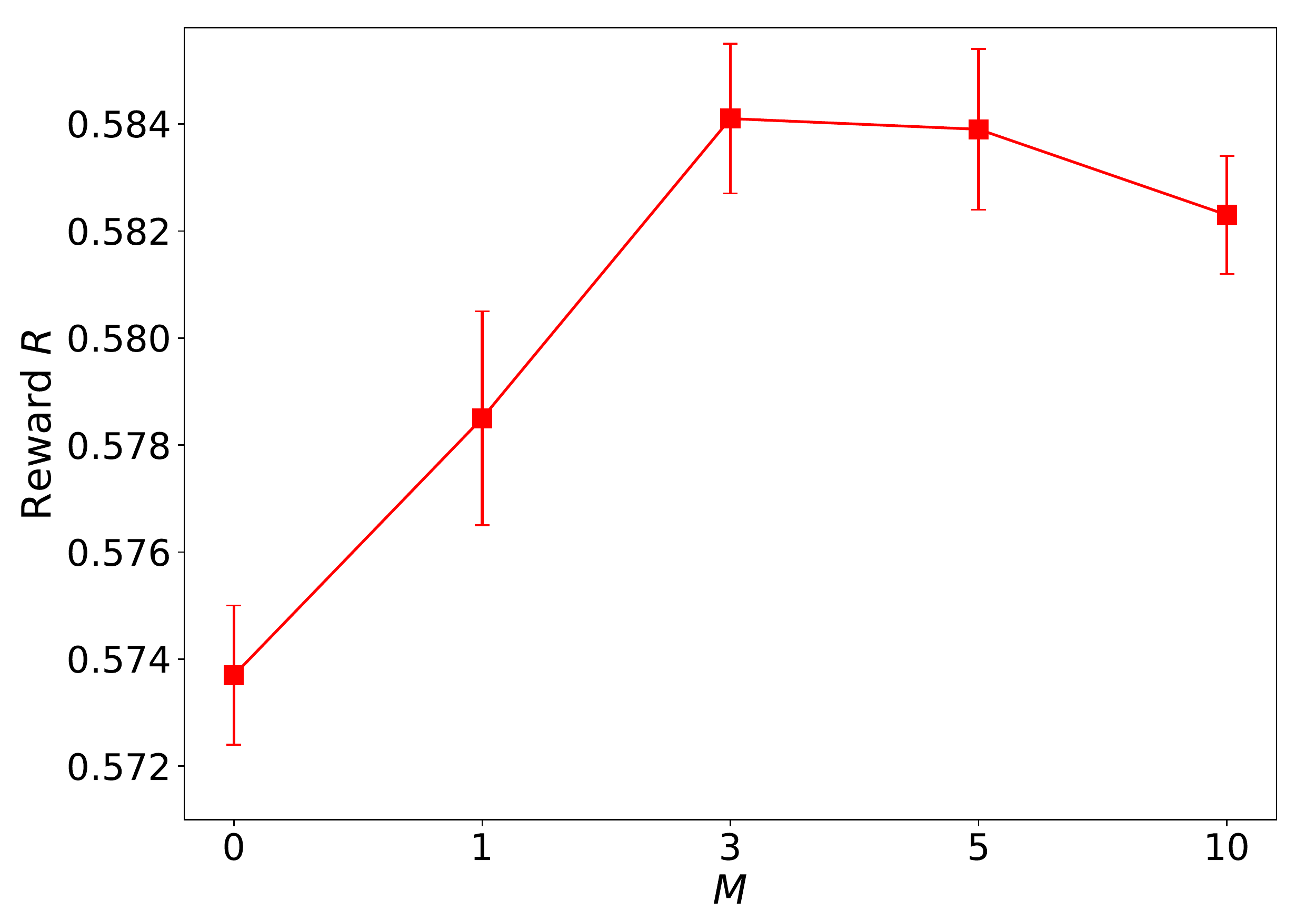} 
    \label{fig:subfig:m} 
  \end{minipage}%
  } 
  \subfigure[{The reward curve of $L$.}]{ 
    \centering
    \begin{minipage}{0.303\linewidth}
    \includegraphics[width=\textwidth]{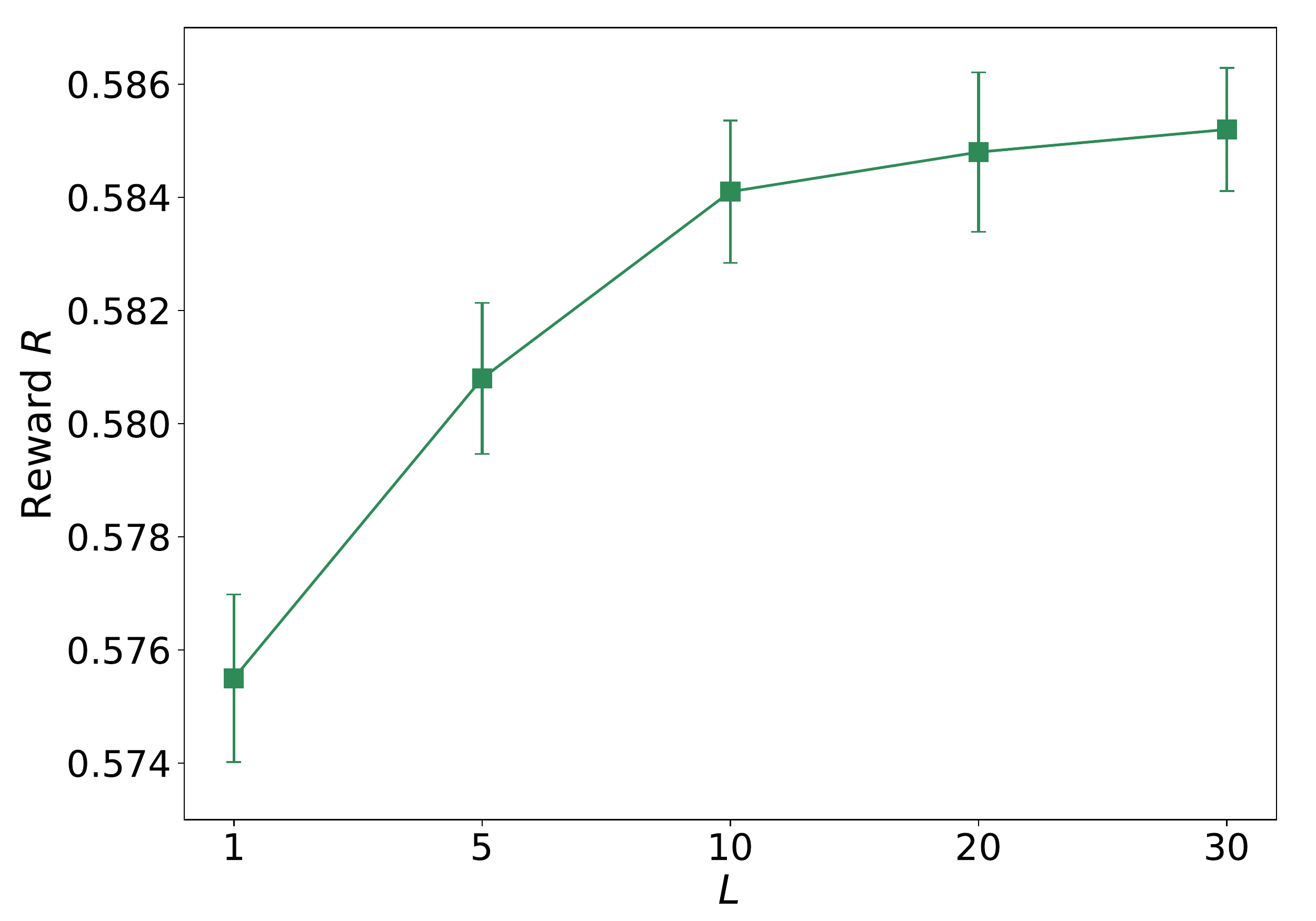}
    \label{fig:subfig:l} 
  \end{minipage}%
  } 
  \subfigure[{The reward curve of $K$.}]{ 
    \centering
    \begin{minipage}{0.303\linewidth}
    \includegraphics[width=\textwidth]{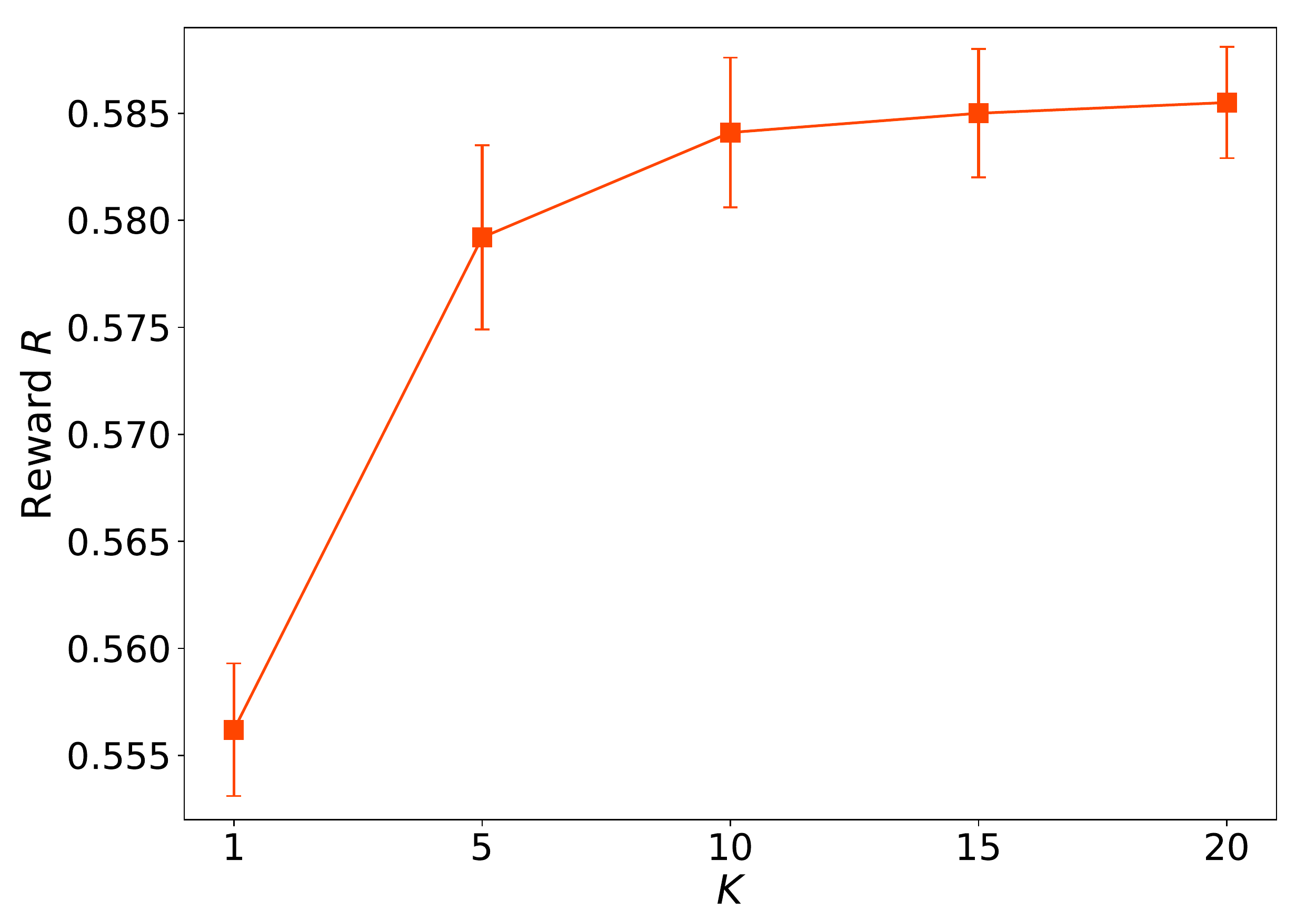} 
    \label{fig:subfig:k} 
  \end{minipage}%
  } 
  \caption{{The experimental results on the sensitivity of $M$ , $L$ and $K$.}
  }
  \label{fig:kml}
\end{figure*}

\begin{table}
  \caption{Compared to $\alpha_2, \alpha_3$ are $0$, the average improvement for four indicators when $\alpha_2$ and $\alpha_3$ change simultaneously.}
  \centering
   
 \subtable[The results of improvement when $\alpha_1$ is 0.0]{
  \resizebox{0.9\linewidth}{14.3mm}{
    \begin{tabular}{|c|ccccc|}
      \hline
      \diagbox[height=14pt]{$\alpha_2$}{$\alpha_3$}  & 0.0 & 0.005 & 0.01 & 0.05  & 0.1   \\ \hline
      0.0 & \cellcolor{o1}-0.76\% & \cellcolor{o1}-0.74\% & \cellcolor{o3}-0.25\% &\cellcolor{o4} -0.02\% & \cellcolor{o5}0.03\% \\ 
      0.005 & \cellcolor{o1}-0.64\% & \cellcolor{o1}-0.72\% & \cellcolor{o5}0.05\% & \cellcolor{o4}-0.04\% & \cellcolor{o3}-0.20\% \\ 
      0.01  &  \cellcolor{o2}-0.43\% &  \cellcolor{o2}-0.43\% &  \cellcolor{o4}-0.09\% &  \cellcolor{o5}0.05\% &  \cellcolor{o4}-0.13\% \\ 
      0.05  &  \cellcolor{o2}-0.51\% &  \cellcolor{o2}-0.57\% &  \cellcolor{o4}-0.09\% &  \cellcolor{o7}0.13\% &  \cellcolor{o3}-0.25\% \\ 
      0.1   &  \cellcolor{o3}-0.25\% &  \cellcolor{o4}-0.13\% &  \cellcolor{o3}-0.25\% &  \cellcolor{o2}-0.46\% &  \cellcolor{o1}-0.72\% \\ \hline
      \end{tabular}
  \label{tb:s1}
  }
 }
   
  \qquad
   
\subtable[The results of improvement when $\alpha_1$ is 0.005]{        
  \resizebox{0.9\linewidth}{14.3mm}{
    \begin{tabular}{|c|ccccc|}
      \hline
      \diagbox[height=14pt]{$\alpha_2$}{$\alpha_3$} & 0.0 & 0.005 & 0.01  & 0.05  & 0.1   \\ \hline
      0.0 &   \cellcolor{o1}-0.32\% &  \cellcolor{o2}-0.02\% &  \cellcolor{o7}0.87\% &  \cellcolor{o8}1.18\% &  \cellcolor{o7}0.85\% \\ 
      0.005 &  \cellcolor{o2}-0.09\% &  \cellcolor{o3}0.20\% &  \cellcolor{o8}1.04\% &  \cellcolor{o7}0.80\% &  \cellcolor{o7}0.76\% \\ 
      0.01  &  \cellcolor{o2}0.05\% &  \cellcolor{o6}0.62\% &  \cellcolor{o8}1.13\% &  \cellcolor{o9}1.56\% &  \cellcolor{o5}0.52\% \\ 
      0.05  &  \cellcolor{o3}0.15\% &  \cellcolor{o9}1.49\% &  \cellcolor{o9}1.58\% &  \cellcolor{o10}1.74\% &  \cellcolor{o4}0.19\% \\ 
      0.1   &  \cellcolor{o1}-0.53\% &  \cellcolor{o4}0.25\% &  \cellcolor{o7}0.83\% &  \cellcolor{o4}0.19\% &  \cellcolor{o1}-0.44\% \\ \hline
      \end{tabular}
  \label{tb:s2}
  }
}

  \qquad
   
  \subtable[The results of improvement when $\alpha_1$ is 0.01]{        
    \resizebox{0.9\linewidth}{14.3mm}{
      \begin{tabular}{|c|ccccc|}
      \hline
      \diagbox[height=14pt]{$\alpha_2$}{$\alpha_3$} & 0.0 & 0.005 & 0.01  & 0.05  & 0.1   \\ \hline
      0.0 & \cellcolor{o1}-0.44\% & \cellcolor{o1}-0.13\% & \cellcolor{o2}-0.08\% & \cellcolor{o2}-0.04\% & \cellcolor{o1}-0.15\% \\ 
      0.005 & \cellcolor{o1}-0.16\% & \cellcolor{o3}0.22\% & \cellcolor{o4}0.85\% & \cellcolor{o7}1.23\% & \cellcolor{o4}0.57\% \\ 
      0.01  & \cellcolor{o2}-0.04\% & \cellcolor{o4}0.64\% & \cellcolor{o5}1.08\% & \cellcolor{o8}1.53\% & \cellcolor{o5}1.09\% \\ 
      0.05  & \cellcolor{o2}0.10\% & \cellcolor{o5}0.95\% & \cellcolor{o7}1.30\% & \cellcolor{o10}1.95\% & \cellcolor{o9}1.81\% \\ 
      0.1   & \cellcolor{o2}-0.01\% & \cellcolor{o4}0.55\% & \cellcolor{o5}1.02\% & \cellcolor{o9}1.76\% & \cellcolor{o7}1.25\% \\ \hline
      \end{tabular}
  \label{tb:s3}
  }
  }

  \qquad
   
 \subtable[The results of improvement when $\alpha_1$ is 0.05]{        
  \resizebox{0.9\linewidth}{14.3mm}{
    \begin{tabular}{|c|ccccc|}
      \hline
      \diagbox[height=14pt]{$\alpha_2$}{$\alpha_3$} & 0.0 & 0.005 & 0.01  & 0.05  & 0.1   \\ \hline
      0.0 & \cellcolor{o1}-0.23\% & \cellcolor{o2}0.03\% & \cellcolor{o8}1.02\% & \cellcolor{o7}0.97\% & \cellcolor{o7}0.92\% \\ 
      0.005 & \cellcolor{o2}0.03\% & \cellcolor{o3}0.25\% & \cellcolor{o8}1.18\% & \cellcolor{o7}0.94\% & \cellcolor{o6}0.87\% \\ 
      0.01  & \cellcolor{o3}0.25\% & \cellcolor{o8}1.30\% & \cellcolor{o8}1.16\% & \cellcolor{o9}1.69\% & \cellcolor{o5}0.62\% \\ 
      0.05  & \cellcolor{o5}0.62\% & \cellcolor{o9}1.65\% & \cellcolor{o9}1.74\% & \cellcolor{o10}1.91\% & \cellcolor{o3}0.22\% \\ 
      0.1   & \cellcolor{o1}-0.46\% & \cellcolor{o4}0.31\% & \cellcolor{o7}0.94\% & \cellcolor{o4}0.34\% & \cellcolor{o1}-0.27\% \\ \hline
      \end{tabular}
  \label{tb:s4}
  }
 }

  \qquad
   
 \subtable[The results of improvement when $\alpha_1$ is 0.1]{   
  \resizebox{0.9\linewidth}{14.3mm}{
    \begin{tabular}{|c|ccccc|}
      \hline
     \diagbox[height=14pt]{$\alpha_2$}{$\alpha_3$} & 0.0 & 0.005 & 0.01  & 0.05  & 0.1   \\ \hline
      0.0 &  \cellcolor{o1}-0.34\% &  \cellcolor{o1}-0.36\% &  \cellcolor{o5}0.57\% &  \cellcolor{o7}0.95\% &  \cellcolor{o6}0.74\% \\ 
      0.005 &  \cellcolor{o2}-0.06\% &  \cellcolor{o2}-0.15\% &  \cellcolor{o7}0.90\% &  \cellcolor{o6}0.78\% &  \cellcolor{o5}0.50\% \\ 
      0.01  &  \cellcolor{o4}0.10\% &  \cellcolor{o7}0.99\% &  \cellcolor{o6}0.73\% &  \cellcolor{o8}1.34\% &  \cellcolor{o4}0.25\% \\ 
      0.05  &  \cellcolor{o8}1.39\% &  \cellcolor{o7}1.09\% & \cellcolor{o9} 1.58\% &  \cellcolor{o8}1.35\% &  \cellcolor{o3}-0.02\% \\ 
      0.1   &  \cellcolor{o1}-0.84\% &  \cellcolor{o4}0.25\% &  \cellcolor{o5}0.55\% &  \cellcolor{o3}-0.06\% &  \cellcolor{o1}-0.48\% \\ \hline
      \end{tabular}
  \label{tb:s5}
  }
  }
  \label{tb:alpha}
  \end{table}

\subsubsection{Performance Comparison}
We keep the percentage of ads exposed at the same level for all methods to ensure comparability \cite{liao2021cross}.
The offline experimental results are shown in Table \ref{tab:all_res} and we have the following observations: 
Intuitively, our method has made great improvements over state-of-the-art baselines in both revenue indicators and experience indicators. And we have the following detailed observations from the experimental results: i) Compared with all RL-based ads allocation baselines, our method achieves strongly competitive performance on both the platform revenue and the user experience. Specifically, our method improves over the best baseline w.r.t. $R^\text{ad}$, $R^\text{fee}$, $R^\text{cxr}$ and $R^\text{ex}$ by 2.38\%, 1.81\%, 1.31\% and 0.73\% separately. 
ii) Compared with different types of auxiliary task methods, our results are substantially better than corresponding types of baseline separately. 
The superior performance of our method justifies that the agent learned more effectively with the help of our designed auxiliary tasks, which can more effectively utilize the side information on ads allocation scenario.

\subsubsection{Ablation Study}
To verify the impact of three auxiliary tasks, we study four ablated variants of our method (i.e., w/o $\text{RAT}$, w/o $\text{PAT}$, w/o $\text{CLAT}$, w/o $\text{all AT}$) and have the following findings:
i) The performance gap between w/ and w/o the first auxiliary task verifies the effectiveness of reconstruction-based auxiliary task, since the key information is embedded in the representation. ii) The performance gap between w/ and w/o the second auxiliary task verifies the effectiveness of prediction-based auxiliary task, which brings in supervised information based on users behaviors to jointly guide the agent in training. iii) The performance gap between w/ and w/o the third auxiliary task verifies the effectiveness of contrastive-learning based auxiliary task, which makes the distinction between different types of state-action pair representations more reasonable. iv) The performance gap between w/ and w/o all auxiliary tasks verifies the fact that the three auxiliary tasks can greatly improve the performance of an agent for ads allocation.

\subsubsection{Hyperparameter Analysis}
We analyze the sensitivity of these four types of hyperparameters: 
\begin{itemize}[leftmargin=*]
  \item \textbf{Coefficients $\alpha_1$, $\alpha_2$, $\alpha_3$}. 
  We perform a detailed analysis of $\alpha_1$, $\alpha_2$, $\alpha_3$ and have the following findings: i) The sensitivity of $\alpha_1$, $\alpha_2$, $\alpha_3$ is shown in Figure \ref{fig:alpha}. Increasing $\alpha_1$, $\alpha_2$ and $\alpha_3$ within a certain range can improve the performance of the agent in auxiliary tasks, which further helps the performance in main task. But if $\alpha_1$, $\alpha_2$ and $\alpha_3$ are too large, it would cause the performance of main task to degrade. Therefore, if the appropriate $\alpha_1$, $\alpha_2$ and $\alpha_3$ are chosen, the auxiliary tasks would greatly improve the performance of the agent in main task. ii) Table \ref{tb:alpha} illustrates the performance when different weights of auxiliary tasks are used simultaneously. The best performance is obtained when $\alpha_1$ is $0.01$, $\alpha_2$ is $0.05$ and $\alpha_3$ is $0.05$. One possible explanation is that different auxiliary tasks can effectively guide the representation to learn in a target direction within a certain range. But if one weight of an auxiliary task is too large, it may cause the learning direction to be dominated by this task, resulting in a decrease in the performance.

  \item \textbf{The amount of information used for reconstruction $M$}. In reconstruction-based auxiliary task, we select top $M$ most concerned factors to build the reconstruction-based auxiliary task. We experiment with $M$ from $1$ to $10$. The experimental result is a typical convex curve and the optimal result is obtained when $M$ is $3$. One reasonable explanation is that key information is helpful to the learning of the representation to a certain extent. But if there is too much information, it will also lead to learning difficulties for the agent.
  
  \item \textbf{The size of contrastive sample set $L$}. In contrastive-learning based auxiliary task, we construct a comparative sample set for each sample, which consists of $1$ positive sample of the same type and $L-1$ randomly sampled negative samples. As shown in figure \ref{fig:subfig:l}, increasing $L$ within a certain range can effectively improve the performance, but the performance will not increase if $L$ is larger than a threshold. From the result it is clear that maintaining a reasonable size $L$ can effectively save computing resources while keeping the performance.
  
    \item \textbf{The number of allocated slots in each request $K$}. As shown in Figure \ref{fig:subfig:k}, 
  increase $K$ can boost the performance. The best performance is obtained when the number of allocated slots in each request is taken as $10$. One reasonable explanation is that the list-wise information increases as $K$ increases. But the action space grows exponentially with $K$. If $K$ is too large, the huge action space would make decision-making more difficult.
  
\end{itemize}




\subsection{Online Results}
We compare our method with Cross DQN and both strategies are deployed on the Meituan platform through online A/B test. We keep total percentage of ads exposed the same for two methods for a fair comparison. The two experimental groups use the same number of users and are observed for two consecutive weeks. As a result, we find that $R^\text{ad}$, $R^\text{fee}$, $R^\text{cxr}$ and $R^\text{ex}$ increase by 2.92\%, 1.91\%, 2.21\% and 1.13\%, which demonstrates that our method can effectively improve both platform revenue and user experience.

\section{Conclusion and future work}
In this paper, we propose three different types of auxiliary tasks to learn an efficient and generalizable representation in the high-dimensional state-action space in the ads allocation scenario. 
Specifically, the three different types of auxiliary tasks are based on reconstruction, prediction, and contrastive learning respectively. 
The reconstruction based auxiliary task helps to learn a representation that embeds the key factors that affect the users.
The prediction based auxiliary task extracts labels based on the  behavior of the users and learns a representation that is predictive of the behavior-based rewards.
The contrastive learning based auxiliary task helps to aggregate semantically similar representations and differentiate different representations.
Practically, both offline experiments and online A/B test have demonstrated the superior performance and efficiency of the proposed method.

However, adding multiple auxiliary tasks at the same time inevitably introduces the challenge that how to balance multiple auxiliary tasks. So, how to automatically balance between multiple auxiliary tasks to maximize the platform revenue is one of our priorities in the future. In addition, it is worth noting that our method follows the offline reinforcement learning paradigm. 
Compared with online reinforcement learning, offline reinforcement learning faces additional challenges (such as the distribution shift problem). The impact of these challenges to the ads allocation problem is also a potential research direction in the future.

\balance
\begin{acks}
  We thank the anonymous reviewers for their constructive suggestions and comments. We also thank Fan Yang, Bingqi Zhu, Hui Niu for helpful discussions.
\end{acks}
\bibliographystyle{ACM-Reference-Format}
\bibliography{sample-base}


\begin{thebibliography}{47}


\ifx \showCODEN    \undefined \def \showCODEN     #1{\unskip}     \fi
\ifx \showDOI      \undefined \def \showDOI       #1{#1}\fi
\ifx \showISBNx    \undefined \def \showISBNx     #1{\unskip}     \fi
\ifx \showISBNxiii \undefined \def \showISBNxiii  #1{\unskip}     \fi
\ifx \showISSN     \undefined \def \showISSN      #1{\unskip}     \fi
\ifx \showLCCN     \undefined \def \showLCCN      #1{\unskip}     \fi
\ifx \shownote     \undefined \def \shownote      #1{#1}          \fi
\ifx \showarticletitle \undefined \def \showarticletitle #1{#1}   \fi
\ifx \showURL      \undefined \def \showURL       {\relax}        \fi
\providecommand\bibfield[2]{#2}
\providecommand\bibinfo[2]{#2}
\providecommand\natexlab[1]{#1}
\providecommand\showeprint[2][]{arXiv:#2}

\bibitem[\protect\citeauthoryear{Anand, Racah, Ozair, Bengio, C{\^o}t{\'e}, and
  Hjelm}{Anand et~al\mbox{.}}{2019}]%
        {anand2019unsupervised}
\bibfield{author}{\bibinfo{person}{Ankesh Anand}, \bibinfo{person}{Evan Racah},
  \bibinfo{person}{Sherjil Ozair}, \bibinfo{person}{Yoshua Bengio},
  \bibinfo{person}{Marc-Alexandre C{\^o}t{\'e}}, {and} \bibinfo{person}{R~Devon
  Hjelm}.} \bibinfo{year}{2019}\natexlab{}.
\newblock \showarticletitle{Unsupervised state representation learning in
  atari}.
\newblock \bibinfo{journal}{\emph{arXiv preprint arXiv:1906.08226}}
  (\bibinfo{year}{2019}).
\newblock


\bibitem[\protect\citeauthoryear{Aytar, Pfaff, Budden, Paine, Wang, and
  de~Freitas}{Aytar et~al\mbox{.}}{2018}]%
        {aytar2018playing}
\bibfield{author}{\bibinfo{person}{Yusuf Aytar}, \bibinfo{person}{Tobias
  Pfaff}, \bibinfo{person}{David Budden}, \bibinfo{person}{Tom~Le Paine},
  \bibinfo{person}{Ziyu Wang}, {and} \bibinfo{person}{Nando de Freitas}.}
  \bibinfo{year}{2018}\natexlab{}.
\newblock \showarticletitle{Playing hard exploration games by watching
  youtube}.
\newblock \bibinfo{journal}{\emph{arXiv preprint arXiv:1805.11592}}
  (\bibinfo{year}{2018}).
\newblock


\bibitem[\protect\citeauthoryear{Carrion, Wang, Nair, Luo, Lei, Lin, Chen, Hu,
  Peng, Bao, and Yan}{Carrion et~al\mbox{.}}{2021}]%
        {Carrion2021BlendingAW}
\bibfield{author}{\bibinfo{person}{Carlos Carrion}, \bibinfo{person}{Zenan
  Wang}, \bibinfo{person}{Harikesh Nair}, \bibinfo{person}{Xianghong Luo},
  \bibinfo{person}{Yulin Lei}, \bibinfo{person}{Xiliang Lin},
  \bibinfo{person}{Wenlong Chen}, \bibinfo{person}{Qiyu Hu},
  \bibinfo{person}{Changping Peng}, \bibinfo{person}{Yongjun Bao}, {and}
  \bibinfo{person}{Weipeng~P. Yan}.} \bibinfo{year}{2021}\natexlab{}.
\newblock \showarticletitle{Blending Advertising with Organic Content in
  E-Commerce: A Virtual Bids Optimization Approach}.
\newblock \bibinfo{journal}{\emph{ArXiv}}  \bibinfo{volume}{abs/2105.13556}
  (\bibinfo{year}{2021}).
\newblock


\bibitem[\protect\citeauthoryear{Dwibedi, Tompson, Lynch, and Sermanet}{Dwibedi
  et~al\mbox{.}}{[n.\,d.]}]%
        {dwibedilearning}
\bibfield{author}{\bibinfo{person}{Debidatta Dwibedi},
  \bibinfo{person}{Jonathan Tompson}, \bibinfo{person}{Corey Lynch}, {and}
  \bibinfo{person}{Pierre Sermanet}.} \bibinfo{year}{[n.\,d.]}\natexlab{}.
\newblock \showarticletitle{Learning actionable representations from visual
  observations. In 2018 IEEE}. In \bibinfo{booktitle}{\emph{RSJ International
  Conference on Intelligent Robots and Systems (IROS)}}.
  \bibinfo{pages}{1577--1584}.
\newblock


\bibitem[\protect\citeauthoryear{Feng, Li, Huang, Liu, Ou, Wang, and Zhu}{Feng
  et~al\mbox{.}}{2018}]%
        {Feng2018LearningTC}
\bibfield{author}{\bibinfo{person}{Jun Feng}, \bibinfo{person}{H. Li},
  \bibinfo{person}{Minlie Huang}, \bibinfo{person}{Shichen Liu},
  \bibinfo{person}{Wenwu Ou}, \bibinfo{person}{Zhirong Wang}, {and}
  \bibinfo{person}{Xiaoyan Zhu}.} \bibinfo{year}{2018}\natexlab{}.
\newblock \showarticletitle{Learning to Collaborate: Multi-Scenario Ranking via
  Multi-Agent Reinforcement Learning}.
\newblock \bibinfo{journal}{\emph{Proceedings of the 2018 World Wide Web
  Conference}} (\bibinfo{year}{2018}).
\newblock


\bibitem[\protect\citeauthoryear{Feng, Gong, Sun, Liu, and Ou}{Feng
  et~al\mbox{.}}{2021}]%
        {Feng2021RevisitRS}
\bibfield{author}{\bibinfo{person}{Yufei Feng}, \bibinfo{person}{Yu Gong},
  \bibinfo{person}{Fei Sun}, \bibinfo{person}{Qingwen Liu}, {and}
  \bibinfo{person}{Wenwu Ou}.} \bibinfo{year}{2021}\natexlab{}.
\newblock \showarticletitle{Revisit Recommender System in the Permutation
  Prospective}.
\newblock \bibinfo{journal}{\emph{ArXiv}}  \bibinfo{volume}{abs/2102.12057}
  (\bibinfo{year}{2021}).
\newblock


\bibitem[\protect\citeauthoryear{Finn, Tan, Duan, Darrell, Levine, and
  Abbeel}{Finn et~al\mbox{.}}{2016}]%
        {finn2016deep}
\bibfield{author}{\bibinfo{person}{Chelsea Finn}, \bibinfo{person}{Xin~Yu Tan},
  \bibinfo{person}{Yan Duan}, \bibinfo{person}{Trevor Darrell},
  \bibinfo{person}{Sergey Levine}, {and} \bibinfo{person}{Pieter Abbeel}.}
  \bibinfo{year}{2016}\natexlab{}.
\newblock \showarticletitle{Deep spatial autoencoders for visuomotor learning}.
  In \bibinfo{booktitle}{\emph{2016 IEEE International Conference on Robotics
  and Automation (ICRA)}}. IEEE, \bibinfo{pages}{512--519}.
\newblock


\bibitem[\protect\citeauthoryear{Fran{\c{c}}ois-Lavet, Bengio, Precup, and
  Pineau}{Fran{\c{c}}ois-Lavet et~al\mbox{.}}{2019}]%
        {franccois2019combined}
\bibfield{author}{\bibinfo{person}{Vincent Fran{\c{c}}ois-Lavet},
  \bibinfo{person}{Yoshua Bengio}, \bibinfo{person}{Doina Precup}, {and}
  \bibinfo{person}{Joelle Pineau}.} \bibinfo{year}{2019}\natexlab{}.
\newblock \showarticletitle{Combined reinforcement learning via abstract
  representations}. In \bibinfo{booktitle}{\emph{Proceedings of the AAAI
  Conference on Artificial Intelligence}}, Vol.~\bibinfo{volume}{33}.
  \bibinfo{pages}{3582--3589}.
\newblock


\bibitem[\protect\citeauthoryear{Ghose and Yang}{Ghose and Yang}{2009}]%
        {Ghose2009AnEA}
\bibfield{author}{\bibinfo{person}{A. Ghose} {and} \bibinfo{person}{Sha Yang}.}
  \bibinfo{year}{2009}\natexlab{}.
\newblock \showarticletitle{An Empirical Analysis of Search Engine Advertising:
  Sponsored Search in Electronic Markets}.
\newblock \bibinfo{journal}{\emph{Manag. Sci.}}  \bibinfo{volume}{55}
  (\bibinfo{year}{2009}), \bibinfo{pages}{1605--1622}.
\newblock


\bibitem[\protect\citeauthoryear{Guo, Pires, Piot, Grill, Altch{\'e}, Munos,
  and Azar}{Guo et~al\mbox{.}}{2020}]%
        {guo2020bootstrap}
\bibfield{author}{\bibinfo{person}{Zhaohan~Daniel Guo},
  \bibinfo{person}{Bernardo~Avila Pires}, \bibinfo{person}{Bilal Piot},
  \bibinfo{person}{Jean-Bastien Grill}, \bibinfo{person}{Florent Altch{\'e}},
  \bibinfo{person}{R{\'e}mi Munos}, {and} \bibinfo{person}{Mohammad~Gheshlaghi
  Azar}.} \bibinfo{year}{2020}\natexlab{}.
\newblock \showarticletitle{Bootstrap latent-predictive representations for
  multitask reinforcement learning}. In \bibinfo{booktitle}{\emph{International
  Conference on Machine Learning}}. PMLR, \bibinfo{pages}{3875--3886}.
\newblock


\bibitem[\protect\citeauthoryear{Ha and Schmidhuber}{Ha and
  Schmidhuber}{2018}]%
        {ha2018recurrent}
\bibfield{author}{\bibinfo{person}{David Ha} {and} \bibinfo{person}{J{\"u}rgen
  Schmidhuber}.} \bibinfo{year}{2018}\natexlab{}.
\newblock \showarticletitle{Recurrent world models facilitate policy
  evolution}.
\newblock \bibinfo{journal}{\emph{Advances in neural information processing
  systems}}  \bibinfo{volume}{31} (\bibinfo{year}{2018}).
\newblock


\bibitem[\protect\citeauthoryear{Hafner, Lillicrap, Ba, and Norouzi}{Hafner
  et~al\mbox{.}}{2019a}]%
        {hafner2019dream}
\bibfield{author}{\bibinfo{person}{Danijar Hafner}, \bibinfo{person}{Timothy
  Lillicrap}, \bibinfo{person}{Jimmy Ba}, {and} \bibinfo{person}{Mohammad
  Norouzi}.} \bibinfo{year}{2019}\natexlab{a}.
\newblock \showarticletitle{Dream to control: Learning behaviors by latent
  imagination}.
\newblock \bibinfo{journal}{\emph{arXiv preprint arXiv:1912.01603}}
  (\bibinfo{year}{2019}).
\newblock


\bibitem[\protect\citeauthoryear{Hafner, Lillicrap, Fischer, Villegas, Ha, Lee,
  and Davidson}{Hafner et~al\mbox{.}}{2019b}]%
        {hafner2019learning}
\bibfield{author}{\bibinfo{person}{Danijar Hafner}, \bibinfo{person}{Timothy
  Lillicrap}, \bibinfo{person}{Ian Fischer}, \bibinfo{person}{Ruben Villegas},
  \bibinfo{person}{David Ha}, \bibinfo{person}{Honglak Lee}, {and}
  \bibinfo{person}{James Davidson}.} \bibinfo{year}{2019}\natexlab{b}.
\newblock \showarticletitle{Learning latent dynamics for planning from pixels}.
  In \bibinfo{booktitle}{\emph{International conference on machine learning}}.
  PMLR, \bibinfo{pages}{2555--2565}.
\newblock


\bibitem[\protect\citeauthoryear{Hessel, Soyer, Espeholt, Czarnecki, Schmitt,
  and van Hasselt}{Hessel et~al\mbox{.}}{2019}]%
        {hessel2019multi}
\bibfield{author}{\bibinfo{person}{Matteo Hessel}, \bibinfo{person}{Hubert
  Soyer}, \bibinfo{person}{Lasse Espeholt}, \bibinfo{person}{Wojciech
  Czarnecki}, \bibinfo{person}{Simon Schmitt}, {and} \bibinfo{person}{Hado van
  Hasselt}.} \bibinfo{year}{2019}\natexlab{}.
\newblock \showarticletitle{Multi-task deep reinforcement learning with
  popart}. In \bibinfo{booktitle}{\emph{Proceedings of the AAAI Conference on
  Artificial Intelligence}}, Vol.~\bibinfo{volume}{33}.
  \bibinfo{pages}{3796--3803}.
\newblock


\bibitem[\protect\citeauthoryear{Jaderberg, Mnih, Czarnecki, Schaul, Leibo,
  Silver, and Kavukcuoglu}{Jaderberg et~al\mbox{.}}{2016}]%
        {jaderberg2016reinforcement}
\bibfield{author}{\bibinfo{person}{Max Jaderberg}, \bibinfo{person}{Volodymyr
  Mnih}, \bibinfo{person}{Wojciech~Marian Czarnecki}, \bibinfo{person}{Tom
  Schaul}, \bibinfo{person}{Joel~Z Leibo}, \bibinfo{person}{David Silver},
  {and} \bibinfo{person}{Koray Kavukcuoglu}.} \bibinfo{year}{2016}\natexlab{}.
\newblock \showarticletitle{Reinforcement learning with unsupervised auxiliary
  tasks}.
\newblock \bibinfo{journal}{\emph{arXiv preprint arXiv:1611.05397}}
  (\bibinfo{year}{2016}).
\newblock


\bibitem[\protect\citeauthoryear{Kingma and Ba}{Kingma and Ba}{2014}]%
        {kingma2014adam}
\bibfield{author}{\bibinfo{person}{Diederik~P Kingma} {and}
  \bibinfo{person}{Jimmy Ba}.} \bibinfo{year}{2014}\natexlab{}.
\newblock \showarticletitle{Adam: A method for stochastic optimization}.
\newblock \bibinfo{journal}{\emph{arXiv preprint arXiv:1412.6980}}
  (\bibinfo{year}{2014}).
\newblock


\bibitem[\protect\citeauthoryear{Koutsopoulos}{Koutsopoulos}{2016}]%
        {koutsopoulos2016optimal}
\bibfield{author}{\bibinfo{person}{Iordanis Koutsopoulos}.}
  \bibinfo{year}{2016}\natexlab{}.
\newblock \showarticletitle{Optimal advertisement allocation in online social
  media feeds}. In \bibinfo{booktitle}{\emph{Proceedings of the 8th ACM
  International Workshop on Hot Topics in Planet-scale mObile computing and
  online Social neTworking}}. \bibinfo{pages}{43--48}.
\newblock


\bibitem[\protect\citeauthoryear{Laskin, Srinivas, and Abbeel}{Laskin
  et~al\mbox{.}}{2020}]%
        {laskin2020curl}
\bibfield{author}{\bibinfo{person}{Michael Laskin}, \bibinfo{person}{Aravind
  Srinivas}, {and} \bibinfo{person}{Pieter Abbeel}.}
  \bibinfo{year}{2020}\natexlab{}.
\newblock \showarticletitle{Curl: Contrastive unsupervised representations for
  reinforcement learning}. In \bibinfo{booktitle}{\emph{International
  Conference on Machine Learning}}. PMLR, \bibinfo{pages}{5639--5650}.
\newblock


\bibitem[\protect\citeauthoryear{Li, Wang, Tong, Tan, Zeng, and Zhuang}{Li
  et~al\mbox{.}}{2020}]%
        {li2020deep}
\bibfield{author}{\bibinfo{person}{Xiang Li}, \bibinfo{person}{Chao Wang},
  \bibinfo{person}{Bin Tong}, \bibinfo{person}{Jiwei Tan},
  \bibinfo{person}{Xiaoyi Zeng}, {and} \bibinfo{person}{Tao Zhuang}.}
  \bibinfo{year}{2020}\natexlab{}.
\newblock \showarticletitle{Deep Time-Aware Item Evolution Network for
  Click-Through Rate Prediction}. In \bibinfo{booktitle}{\emph{Proceedings of
  the 29th ACM International Conference on Information \& Knowledge
  Management}}. \bibinfo{pages}{785--794}.
\newblock


\bibitem[\protect\citeauthoryear{Liao, Shi, Wang, Wu, Zhang, Wang, Wang, and
  Wang}{Liao et~al\mbox{.}}{2022}]%
        {liao2022deep}
\bibfield{author}{\bibinfo{person}{Guogang Liao}, \bibinfo{person}{Xiaowen
  Shi}, \bibinfo{person}{Ze Wang}, \bibinfo{person}{Xiaoxu Wu},
  \bibinfo{person}{Chuheng Zhang}, \bibinfo{person}{Yongkang Wang},
  \bibinfo{person}{Xingxing Wang}, {and} \bibinfo{person}{Dong Wang}.}
  \bibinfo{year}{2022}\natexlab{}.
\newblock \showarticletitle{Deep Page-Level Interest Network in Reinforcement
  Learning for Ads Allocation}.
\newblock \bibinfo{journal}{\emph{arXiv preprint arXiv:2204.00377}}
  (\bibinfo{year}{2022}).
\newblock


\bibitem[\protect\citeauthoryear{Liao, Wang, Wu, Shi, Zhang, Wang, Wang, and
  Wang}{Liao et~al\mbox{.}}{2021}]%
        {liao2021cross}
\bibfield{author}{\bibinfo{person}{Guogang Liao}, \bibinfo{person}{Ze Wang},
  \bibinfo{person}{Xiaoxu Wu}, \bibinfo{person}{Xiaowen Shi},
  \bibinfo{person}{Chuheng Zhang}, \bibinfo{person}{Yongkang Wang},
  \bibinfo{person}{Xingxing Wang}, {and} \bibinfo{person}{Dong Wang}.}
  \bibinfo{year}{2021}\natexlab{}.
\newblock \showarticletitle{Cross DQN: Cross Deep Q Network for Ads Allocation
  in Feed}.
\newblock \bibinfo{journal}{\emph{arXiv preprint arXiv:2109.04353}}
  (\bibinfo{year}{2021}).
\newblock


\bibitem[\protect\citeauthoryear{Liu, Zhang, Zhao, Qin, Zhu, Li, Yu, and
  Liu}{Liu et~al\mbox{.}}{2021}]%
        {rcrl}
\bibfield{author}{\bibinfo{person}{Guoqing Liu}, \bibinfo{person}{Chuheng
  Zhang}, \bibinfo{person}{Li Zhao}, \bibinfo{person}{Tao Qin},
  \bibinfo{person}{Jinhua Zhu}, \bibinfo{person}{Jian Li},
  \bibinfo{person}{Nenghai Yu}, {and} \bibinfo{person}{Tie-Yan Liu}.}
  \bibinfo{year}{2021}\natexlab{}.
\newblock \showarticletitle{Return-based contrastive representation learning
  for reinforcement learning}.
\newblock \bibinfo{journal}{\emph{arXiv preprint arXiv:2102.10960}}
  (\bibinfo{year}{2021}).
\newblock


\bibitem[\protect\citeauthoryear{Mazoure, Combes, Doan, Bachman, and
  Hjelm}{Mazoure et~al\mbox{.}}{2020}]%
        {mazoure2020deep}
\bibfield{author}{\bibinfo{person}{Bogdan Mazoure}, \bibinfo{person}{Remi
  Tachet~des Combes}, \bibinfo{person}{Thang Doan}, \bibinfo{person}{Philip
  Bachman}, {and} \bibinfo{person}{R~Devon Hjelm}.}
  \bibinfo{year}{2020}\natexlab{}.
\newblock \showarticletitle{Deep reinforcement and infomax learning}.
\newblock \bibinfo{journal}{\emph{arXiv preprint arXiv:2006.07217}}
  (\bibinfo{year}{2020}).
\newblock


\bibitem[\protect\citeauthoryear{Mehta}{Mehta}{2013}]%
        {Mehta2013OnlineMA}
\bibfield{author}{\bibinfo{person}{Aranyak Mehta}.}
  \bibinfo{year}{2013}\natexlab{}.
\newblock \showarticletitle{Online Matching and Ad Allocation}.
\newblock \bibinfo{journal}{\emph{Found. Trends Theor. Comput. Sci.}}
  \bibinfo{volume}{8} (\bibinfo{year}{2013}), \bibinfo{pages}{265--368}.
\newblock


\bibitem[\protect\citeauthoryear{Mirowski, Pascanu, Viola, Soyer, Ballard,
  Banino, Denil, Goroshin, Sifre, Kavukcuoglu, et~al\mbox{.}}{Mirowski
  et~al\mbox{.}}{2016}]%
        {mirowski2016learning}
\bibfield{author}{\bibinfo{person}{Piotr Mirowski}, \bibinfo{person}{Razvan
  Pascanu}, \bibinfo{person}{Fabio Viola}, \bibinfo{person}{Hubert Soyer},
  \bibinfo{person}{Andrew~J Ballard}, \bibinfo{person}{Andrea Banino},
  \bibinfo{person}{Misha Denil}, \bibinfo{person}{Ross Goroshin},
  \bibinfo{person}{Laurent Sifre}, \bibinfo{person}{Koray Kavukcuoglu},
  {et~al\mbox{.}}} \bibinfo{year}{2016}\natexlab{}.
\newblock \showarticletitle{Learning to navigate in complex environments}.
\newblock \bibinfo{journal}{\emph{arXiv preprint arXiv:1611.03673}}
  (\bibinfo{year}{2016}).
\newblock


\bibitem[\protect\citeauthoryear{Mnih, Kavukcuoglu, Silver, Rusu, Veness,
  Bellemare, Graves, Riedmiller, Fidjeland, Ostrovski, et~al\mbox{.}}{Mnih
  et~al\mbox{.}}{2015}]%
        {mnih2015human}
\bibfield{author}{\bibinfo{person}{Volodymyr Mnih}, \bibinfo{person}{Koray
  Kavukcuoglu}, \bibinfo{person}{David Silver}, \bibinfo{person}{Andrei~A
  Rusu}, \bibinfo{person}{Joel Veness}, \bibinfo{person}{Marc~G Bellemare},
  \bibinfo{person}{Alex Graves}, \bibinfo{person}{Martin Riedmiller},
  \bibinfo{person}{Andreas~K Fidjeland}, \bibinfo{person}{Georg Ostrovski},
  {et~al\mbox{.}}} \bibinfo{year}{2015}\natexlab{}.
\newblock \showarticletitle{Human-level control through deep reinforcement
  learning}.
\newblock \bibinfo{journal}{\emph{nature}} \bibinfo{volume}{518},
  \bibinfo{number}{7540} (\bibinfo{year}{2015}), \bibinfo{pages}{529--533}.
\newblock


\bibitem[\protect\citeauthoryear{Ni, Huang, Cheng, and Gao}{Ni
  et~al\mbox{.}}{2021}]%
        {ni2021effective}
\bibfield{author}{\bibinfo{person}{Juan Ni}, \bibinfo{person}{Zhenhua Huang},
  \bibinfo{person}{Jiujun Cheng}, {and} \bibinfo{person}{Shangce Gao}.}
  \bibinfo{year}{2021}\natexlab{}.
\newblock \showarticletitle{An effective recommendation model based on deep
  representation learning}.
\newblock \bibinfo{journal}{\emph{Information Sciences}}  \bibinfo{volume}{542}
  (\bibinfo{year}{2021}), \bibinfo{pages}{324--342}.
\newblock


\bibitem[\protect\citeauthoryear{Ouyang, Zhang, Ren, Qi, Liu, and Du}{Ouyang
  et~al\mbox{.}}{2019}]%
        {ouyang2019representation}
\bibfield{author}{\bibinfo{person}{Wentao Ouyang}, \bibinfo{person}{Xiuwu
  Zhang}, \bibinfo{person}{Shukui Ren}, \bibinfo{person}{Chao Qi},
  \bibinfo{person}{Zhaojie Liu}, {and} \bibinfo{person}{Yanlong Du}.}
  \bibinfo{year}{2019}\natexlab{}.
\newblock \showarticletitle{Representation learning-assisted click-through rate
  prediction}.
\newblock \bibinfo{journal}{\emph{arXiv preprint arXiv:1906.04365}}
  (\bibinfo{year}{2019}).
\newblock


\bibitem[\protect\citeauthoryear{Ouyang, Zhang, Zhao, Luo, Zhang, Zou, Liu, and
  Du}{Ouyang et~al\mbox{.}}{2020}]%
        {ouyang2020minet}
\bibfield{author}{\bibinfo{person}{Wentao Ouyang}, \bibinfo{person}{Xiuwu
  Zhang}, \bibinfo{person}{Lei Zhao}, \bibinfo{person}{Jinmei Luo},
  \bibinfo{person}{Yu Zhang}, \bibinfo{person}{Heng Zou},
  \bibinfo{person}{Zhaojie Liu}, {and} \bibinfo{person}{Yanlong Du}.}
  \bibinfo{year}{2020}\natexlab{}.
\newblock \showarticletitle{MiNet: Mixed Interest Network for Cross-Domain
  Click-Through Rate Prediction}. In \bibinfo{booktitle}{\emph{Proceedings of
  the 29th ACM International Conference on Information \& Knowledge
  Management}}. \bibinfo{pages}{2669--2676}.
\newblock


\bibitem[\protect\citeauthoryear{Sermanet, Lynch, Chebotar, Hsu, Jang, Schaal,
  Levine, and Brain}{Sermanet et~al\mbox{.}}{2018}]%
        {sermanet2018time}
\bibfield{author}{\bibinfo{person}{Pierre Sermanet}, \bibinfo{person}{Corey
  Lynch}, \bibinfo{person}{Yevgen Chebotar}, \bibinfo{person}{Jasmine Hsu},
  \bibinfo{person}{Eric Jang}, \bibinfo{person}{Stefan Schaal},
  \bibinfo{person}{Sergey Levine}, {and} \bibinfo{person}{Google Brain}.}
  \bibinfo{year}{2018}\natexlab{}.
\newblock \showarticletitle{Time-contrastive networks: Self-supervised learning
  from video}. In \bibinfo{booktitle}{\emph{2018 IEEE international conference
  on robotics and automation (ICRA)}}. IEEE, \bibinfo{pages}{1134--1141}.
\newblock


\bibitem[\protect\citeauthoryear{Shelhamer, Mahmoudieh, Argus, and
  Darrell}{Shelhamer et~al\mbox{.}}{2016}]%
        {shelhamer2016loss}
\bibfield{author}{\bibinfo{person}{Evan Shelhamer}, \bibinfo{person}{Parsa
  Mahmoudieh}, \bibinfo{person}{Max Argus}, {and} \bibinfo{person}{Trevor
  Darrell}.} \bibinfo{year}{2016}\natexlab{}.
\newblock \showarticletitle{Loss is its own reward: Self-supervision for
  reinforcement learning}.
\newblock \bibinfo{journal}{\emph{arXiv preprint arXiv:1612.07307}}
  (\bibinfo{year}{2016}).
\newblock


\bibitem[\protect\citeauthoryear{Sohn}{Sohn}{2016}]%
        {sohn2016improved}
\bibfield{author}{\bibinfo{person}{Kihyuk Sohn}.}
  \bibinfo{year}{2016}\natexlab{}.
\newblock \showarticletitle{Improved deep metric learning with multi-class
  n-pair loss objective}.
\newblock \bibinfo{journal}{\emph{Advances in neural information processing
  systems}}  \bibinfo{volume}{29} (\bibinfo{year}{2016}).
\newblock


\bibitem[\protect\citeauthoryear{Sutton, Barto, et~al\mbox{.}}{Sutton
  et~al\mbox{.}}{1998}]%
        {sutton1998introduction}
\bibfield{author}{\bibinfo{person}{Richard~S Sutton}, \bibinfo{person}{Andrew~G
  Barto}, {et~al\mbox{.}}} \bibinfo{year}{1998}\natexlab{}.
\newblock \bibinfo{booktitle}{\emph{Introduction to reinforcement learning}}.
  Vol.~\bibinfo{volume}{135}.
\newblock \bibinfo{publisher}{MIT press Cambridge}.
\newblock


\bibitem[\protect\citeauthoryear{Van~den Oord, Li, and Vinyals}{Van~den Oord
  et~al\mbox{.}}{2018}]%
        {van2018representation}
\bibfield{author}{\bibinfo{person}{Aaron Van~den Oord}, \bibinfo{person}{Yazhe
  Li}, {and} \bibinfo{person}{Oriol Vinyals}.} \bibinfo{year}{2018}\natexlab{}.
\newblock \showarticletitle{Representation learning with contrastive predictive
  coding}.
\newblock \bibinfo{journal}{\emph{arXiv e-prints}} (\bibinfo{year}{2018}),
  \bibinfo{pages}{arXiv--1807}.
\newblock


\bibitem[\protect\citeauthoryear{van~der Pol, Worrall, van Hoof, Oliehoek, and
  Welling}{van~der Pol et~al\mbox{.}}{2020}]%
        {van2020mdp}
\bibfield{author}{\bibinfo{person}{Elise van~der Pol}, \bibinfo{person}{Daniel
  Worrall}, \bibinfo{person}{Herke van Hoof}, \bibinfo{person}{Frans Oliehoek},
  {and} \bibinfo{person}{Max Welling}.} \bibinfo{year}{2020}\natexlab{}.
\newblock \showarticletitle{MDP homomorphic networks: Group symmetries in
  reinforcement learning}.
\newblock \bibinfo{journal}{\emph{Advances in Neural Information Processing
  Systems}}  \bibinfo{volume}{33} (\bibinfo{year}{2020}).
\newblock


\bibitem[\protect\citeauthoryear{Vaswani, Shazeer, Parmar, Uszkoreit, Jones,
  Gomez, Kaiser, and Polosukhin}{Vaswani et~al\mbox{.}}{2017}]%
        {vaswani2017attention}
\bibfield{author}{\bibinfo{person}{Ashish Vaswani}, \bibinfo{person}{Noam
  Shazeer}, \bibinfo{person}{Niki Parmar}, \bibinfo{person}{Jakob Uszkoreit},
  \bibinfo{person}{Llion Jones}, \bibinfo{person}{Aidan~N Gomez},
  \bibinfo{person}{Lukasz Kaiser}, {and} \bibinfo{person}{Illia Polosukhin}.}
  \bibinfo{year}{2017}\natexlab{}.
\newblock \showarticletitle{Attention is all you need}.
\newblock \bibinfo{journal}{\emph{arXiv preprint arXiv:1706.03762}}
  (\bibinfo{year}{2017}).
\newblock


\bibitem[\protect\citeauthoryear{Wang, Li, Tang, Zhang, Chen, and Ru}{Wang
  et~al\mbox{.}}{2011}]%
        {Wang2011LearningTA}
\bibfield{author}{\bibinfo{person}{B. Wang}, \bibinfo{person}{Zhaonan Li},
  \bibinfo{person}{Jie Tang}, \bibinfo{person}{Kuo Zhang},
  \bibinfo{person}{Songcan Chen}, {and} \bibinfo{person}{Liyun Ru}.}
  \bibinfo{year}{2011}\natexlab{}.
\newblock \showarticletitle{Learning to Advertise: How Many Ads Are Enough?}.
  In \bibinfo{booktitle}{\emph{PAKDD}}.
\newblock


\bibitem[\protect\citeauthoryear{Wang, Schaul, Hessel, Hasselt, Lanctot, and
  Freitas}{Wang et~al\mbox{.}}{2016}]%
        {wang2016duelingDQN}
\bibfield{author}{\bibinfo{person}{Ziyu Wang}, \bibinfo{person}{Tom Schaul},
  \bibinfo{person}{Matteo Hessel}, \bibinfo{person}{Hado Hasselt},
  \bibinfo{person}{Marc Lanctot}, {and} \bibinfo{person}{Nando Freitas}.}
  \bibinfo{year}{2016}\natexlab{}.
\newblock \showarticletitle{Dueling network architectures for deep
  reinforcement learning}. In \bibinfo{booktitle}{\emph{International
  conference on machine learning}}. PMLR, \bibinfo{pages}{1995--2003}.
\newblock


\bibitem[\protect\citeauthoryear{Wei, Zeng, Wu, Guo, Hua, and Cai}{Wei
  et~al\mbox{.}}{2020}]%
        {Wei2020GeneratorAC}
\bibfield{author}{\bibinfo{person}{Jianxiong Wei}, \bibinfo{person}{Anxiang
  Zeng}, \bibinfo{person}{Yueqiu Wu}, \bibinfo{person}{Pengxin Guo},
  \bibinfo{person}{Q. Hua}, {and} \bibinfo{person}{Qingpeng Cai}.}
  \bibinfo{year}{2020}\natexlab{}.
\newblock \showarticletitle{Generator and Critic: A Deep Reinforcement Learning
  Approach for Slate Re-ranking in E-commerce}.
\newblock \bibinfo{journal}{\emph{ArXiv}}  \bibinfo{volume}{abs/2005.12206}
  (\bibinfo{year}{2020}).
\newblock


\bibitem[\protect\citeauthoryear{Xie, Zhang, Wang, Xia, and Lin}{Xie
  et~al\mbox{.}}{2021}]%
        {xie2021hierarchical}
\bibfield{author}{\bibinfo{person}{Ruobing Xie}, \bibinfo{person}{Shaoliang
  Zhang}, \bibinfo{person}{Rui Wang}, \bibinfo{person}{Feng Xia}, {and}
  \bibinfo{person}{Leyu Lin}.} \bibinfo{year}{2021}\natexlab{}.
\newblock \showarticletitle{Hierarchical Reinforcement Learning for Integrated
  Recommendation}. In \bibinfo{booktitle}{\emph{Proceedings of AAAI}}.
\newblock


\bibitem[\protect\citeauthoryear{Yan, Xu, Tiwana, and Chatterjee}{Yan
  et~al\mbox{.}}{2020}]%
        {yan2020LinkedInGEA}
\bibfield{author}{\bibinfo{person}{Jinyun Yan}, \bibinfo{person}{Zhiyuan Xu},
  \bibinfo{person}{Birjodh Tiwana}, {and} \bibinfo{person}{Shaunak
  Chatterjee}.} \bibinfo{year}{2020}\natexlab{}.
\newblock \showarticletitle{Ads Allocation in Feed via Constrained
  Optimization}. In \bibinfo{booktitle}{\emph{Proceedings of the 26th ACM
  SIGKDD International Conference on Knowledge Discovery \& Data Mining}}.
  \bibinfo{pages}{3386--3394}.
\newblock


\bibitem[\protect\citeauthoryear{Zhang, Wei, Meng, Hu, and Wang}{Zhang
  et~al\mbox{.}}{2018}]%
        {zhang2018whole}
\bibfield{author}{\bibinfo{person}{Weiru Zhang}, \bibinfo{person}{Chao Wei},
  \bibinfo{person}{Xiaonan Meng}, \bibinfo{person}{Yi Hu}, {and}
  \bibinfo{person}{Hao Wang}.} \bibinfo{year}{2018}\natexlab{}.
\newblock \showarticletitle{The whole-page optimization via dynamic ad
  allocation}. In \bibinfo{booktitle}{\emph{Companion Proceedings of the The
  Web Conference}}. \bibinfo{pages}{1407--1411}.
\newblock


\bibitem[\protect\citeauthoryear{Zhao, Li, An, Lu, Yang, and Chu}{Zhao
  et~al\mbox{.}}{2018}]%
        {Zhao2018ImpressionAF}
\bibfield{author}{\bibinfo{person}{Mengchen Zhao}, \bibinfo{person}{Z. Li},
  \bibinfo{person}{Bo An}, \bibinfo{person}{Haifeng Lu}, \bibinfo{person}{Yifan
  Yang}, {and} \bibinfo{person}{Chen Chu}.} \bibinfo{year}{2018}\natexlab{}.
\newblock \showarticletitle{Impression Allocation for Combating Fraud in
  E-commerce Via Deep Reinforcement Learning with Action Norm Penalty}. In
  \bibinfo{booktitle}{\emph{IJCAI}}.
\newblock


\bibitem[\protect\citeauthoryear{Zhao, Gu, Zhang, Yang, Liu, Liu, and
  Tang}{Zhao et~al\mbox{.}}{2021}]%
        {zhao2019deep}
\bibfield{author}{\bibinfo{person}{Xiangyu Zhao}, \bibinfo{person}{Changsheng
  Gu}, \bibinfo{person}{Haoshenglun Zhang}, \bibinfo{person}{Xiwang Yang},
  \bibinfo{person}{Xiaobing Liu}, \bibinfo{person}{Hui Liu}, {and}
  \bibinfo{person}{Jiliang Tang}.} \bibinfo{year}{2021}\natexlab{}.
\newblock \showarticletitle{DEAR: Deep Reinforcement Learning for Online
  Advertising Impression in Recommender Systems}. In
  \bibinfo{booktitle}{\emph{Proceedings of the AAAI Conference on Artificial
  Intelligence}}, Vol.~\bibinfo{volume}{35}. \bibinfo{pages}{750--758}.
\newblock


\bibitem[\protect\citeauthoryear{Zhao, Zheng, Yang, Liu, and Tang}{Zhao
  et~al\mbox{.}}{2020}]%
        {zhao2020jointly}
\bibfield{author}{\bibinfo{person}{Xiangyu Zhao}, \bibinfo{person}{Xudong
  Zheng}, \bibinfo{person}{Xiwang Yang}, \bibinfo{person}{Xiaobing Liu}, {and}
  \bibinfo{person}{Jiliang Tang}.} \bibinfo{year}{2020}\natexlab{}.
\newblock \showarticletitle{Jointly learning to recommend and advertise}. In
  \bibinfo{booktitle}{\emph{Proceedings of the 26th ACM SIGKDD International
  Conference on Knowledge Discovery \& Data Mining}}.
  \bibinfo{pages}{3319--3327}.
\newblock


\bibitem[\protect\citeauthoryear{Zhou, Zhu, Song, Fan, Zhu, Ma, Yan, Jin, Li,
  and Gai}{Zhou et~al\mbox{.}}{2018}]%
        {zhou2018DIN}
\bibfield{author}{\bibinfo{person}{Guorui Zhou}, \bibinfo{person}{Xiaoqiang
  Zhu}, \bibinfo{person}{Chenru Song}, \bibinfo{person}{Ying Fan},
  \bibinfo{person}{Han Zhu}, \bibinfo{person}{Xiao Ma},
  \bibinfo{person}{Yanghui Yan}, \bibinfo{person}{Junqi Jin},
  \bibinfo{person}{Han Li}, {and} \bibinfo{person}{Kun Gai}.}
  \bibinfo{year}{2018}\natexlab{}.
\newblock \showarticletitle{Deep interest network for click-through rate
  prediction}. In \bibinfo{booktitle}{\emph{Proceedings of the 24th ACM SIGKDD
  International Conference on Knowledge Discovery \& Data Mining}}.
  \bibinfo{pages}{1059--1068}.
\newblock


\bibitem[\protect\citeauthoryear{Zhu, Wu, Qiang, Yuan, and Li}{Zhu
  et~al\mbox{.}}{2021}]%
        {zhu2021representation}
\bibfield{author}{\bibinfo{person}{Yi Zhu}, \bibinfo{person}{Xindong Wu},
  \bibinfo{person}{Jipeng Qiang}, \bibinfo{person}{Yunhao Yuan}, {and}
  \bibinfo{person}{Yun Li}.} \bibinfo{year}{2021}\natexlab{}.
\newblock \showarticletitle{Representation learning with collaborative
  autoencoder for personalized recommendation}.
\newblock \bibinfo{journal}{\emph{Expert Systems with Applications}}
  \bibinfo{volume}{186} (\bibinfo{year}{2021}), \bibinfo{pages}{115825}.
\newblock


\end{thebibliography}

\appendix

\end{document}